%% file: 0_main.tex
\documentclass[conference]{IEEEtran}
\IEEEoverridecommandlockouts

\usepackage{cite}
\usepackage{amsmath,amssymb,amsfonts}
\usepackage{algorithmic}
\usepackage{graphicx}
\usepackage{textcomp}
\usepackage{xcolor}
\usepackage{amsthm}
\usepackage{etoolbox}
\usepackage{array}
\usepackage{graphicx}
\usepackage{xcolor}
\usepackage{threeparttable}
\usepackage{subcaption}
\usepackage{flushend}

\usepackage{bm}
\usepackage{booktabs}
\usepackage{xspace}
\usepackage{threeparttable}
\usepackage{multirow}
\usepackage{xspace}
\usepackage{dsfont}
\usepackage{epsfig}
\usepackage{epstopdf}
\usepackage{booktabs}
\usepackage{bm}
\usepackage{listings}
\usepackage{tabu}
\usepackage{longtable}
\usepackage{threeparttable}
\usepackage{marvosym}
\usepackage{bigstrut}
\usepackage{hyperref}
\usepackage{soul}
\usepackage{float}
\usepackage{tabularx}
\usepackage{enumitem}
\usepackage[linesnumbered,ruled,vlined, boxed]{algorithm2e}
\usepackage{diagbox}
\usepackage[normalem]{ulem}
\useunder{\uline}{\ul}{}
\usepackage{colortbl}
\newcommand{\ie}{\emph{i.e., }}

\newcommand{\eg}{\emph{e.g., }}

\newcommand{\etc}{\emph{etc}}

\newcommand{\paratitle}[1]{\vspace{1.2ex}\noindent\textbf{#1}}
\newcommand{\eat}[1]{}

\newcommand{\name}{{\Name}\xspace}
\newcommand{\Name}{{BIGCity}}
\newcommand{\Sname}{BIG}
\newcommand{\submodel}{VMTP\xspace}
\newcommand{\backbone}{GPT-2}
\newcommand{\bj}{{BJ}\xspace}
\newcommand{\xa}{{XA}\xspace}
\newcommand{\cd}{{CD}\xspace}

\def\BibTeX{{\rm B\kern-.05em{\sc i\kern-.025em b}\kern-.08em
    T\kern-.1667em\lower.7ex\hbox{E}\kern-.125emX}}

\newtheorem{mydef}{Definition}

\begin{document}

% \title{\name: A Versatile Model for Unified Multi-level Spatiotemporal Data Analysis}
\title{\name: A Universal Spatiotemporal Model for Unified Trajectory and Traffic State Data Analysis}

\author{
    \IEEEauthorblockN{Xie Yu\textsuperscript{$\dagger$}, Jingyuan Wang\textsuperscript{$\dagger$$\ast$}\thanks{$^{\ast}$ Corresponding Author: Jingyuan Wang}, Yifan Yang\textsuperscript{$\dagger$}, Qian Huang\textsuperscript{$\ddagger$}, Ke Qu\textsuperscript{$\ddagger$}}
    \IEEEauthorblockA{{$\dagger$} School of Computer Science and Engineering, Beihang University, Beijing, China \\ {$\ddagger$} Huawei Technologies Co., Ltd, Beijing, China }
   \IEEEauthorblockA{\{yuxie\_scse, jywang, yfyang\}@buaa.edu.cn; \{huangqian16, quke\}@huawei.com}
}

\maketitle

\begin{abstract}
Spatiotemporal (ST) data analysis is a critical area of research in data engineering. Typical dynamic ST data includes trajectory data (representing individual-level mobility) and traffic state data (representing population-level mobility). Traditional studies often treat trajectory and traffic state data as distinct, independent modalities, each tailored to specific tasks within a single modality. However, real-world applications, such as navigation apps, require joint analysis of trajectory and traffic state data. Treating these data types as two separate domains can lead to suboptimal model performance. Although recent advances in ST data pre-training and ST foundation models aim to develop universal models for ST data analysis, most existing models are ``multi-task, solo-data modality'' (MTSM), meaning they can handle multiple tasks within either trajectory data or traffic state data, but not both simultaneously.

To address this gap, this paper introduces \name, the first multi-task, multi-data modality (MTMD) model for ST data analysis. The model targets two key challenges in designing an MTMD ST model: (1) unifying the representations of different ST data modalities, and (2) unifying heterogeneous ST analysis tasks. To overcome the first challenge, \name introduces a novel ST-unit that represents both trajectories and traffic states in a unified format. Additionally, for the second challenge, \name adopts a tunable large model with ST task-oriented prompt, enabling it to perform a range of heterogeneous tasks without the need for fine-tuning. Extensive experiments on real-world datasets demonstrate that \name achieves state-of-the-art performance across 8 tasks, outperforming 18 baselines. To the best of our knowledge, \name is the first model capable of handling both trajectories and traffic states for diverse heterogeneous tasks. Our code are available at \url{https://github.com/bigscity/BIGCity}.
\end{abstract}


\input{content/wang_version/1_introduction}

\input{content/6_related_work}

\input{content/wang_version/2_pre_wang}

\input{content/wang_version/3_2_tokenizer}

\input{content/wang_version/3_3_backbone_model}

\input{content/wang_version/3_4_training}

\input{content/4_experiments}

\input{content/5_model_analysis}

\vspace{0.2cm}
\section{Conclusions}
\vspace{0.25cm}
This study introduced the \name model, a foundation model for urban spatiotemporal learning. By employing unified ST representations and textual prompts, \name streamlined the processing of diverse spatiotemporal data and tasks within a single framework. Extensive experiments demonstrated \name's versatility and robustness. In addition, Although \name has a large volume of parameters, it still demonstrates reasonable efficiency and impressive scalability. {\em Future work}: The current \name model focused solely on road segments, excluding other spatial elements such as POIs and grids. We consider incorporating these elements is a promising direction for future research.

% \newpage

\bibliographystyle{IEEEtran}
\bibliography{0_main}

\flushend

\end{document}

%% file: content/wang_version/1_introduction.tex
\section{Introduction}

Spatiotemporal (ST) data analysis has been a cornerstone of research in data engineering, with models such as trajectory analysis and traffic state prediction playing critical roles in various domains. These models are integral to the development of intelligent transportation systems (ITS)~\cite{hoteit2014estimating, ji2022precision}, smart cities~\cite{chen2021robust, li2022spatial, hettige2024airphynet, wang2021dgeye, wang2018inferring, wang2017no}, and location-based service (LBS) applications~\cite{gao2023predicting}, enabling advanced solutions for urban mobility, infrastructure planning, and personalized services.

From a data perspective, ST data analysis models are categorized into two types based on the targeted data: individual-level models for trajectory data and population-level models for traffic state data. Individual-level trajectory models, such as next-hop prediction~\cite{trajectory2vec, Toast, JCLRNT, ma2024more}, trajectory-user linkage~\cite{Trembr, jiang2023self}, and trajectory traffic pattern classification~\cite{START}, process mobility trajectories of individuals on road networks or among POIs to uncover patterns in individual human mobility. Population-level traffic state models, including traffic state prediction~\cite{wu2020connecting, li2021traffic}, traffic state imputation~\cite{deng2021st, guo2023self} and \etc, analyze time series data on traffic metrics like speed, density, and flow volumes. These models aim to capture spatiotemporal correlations at a population level, reflecting crowd behaviors in transportation systems, such as urban road networks.

\begin{figure}[t]
     \centering
     \includegraphics[width=1.0\columnwidth, page=1]{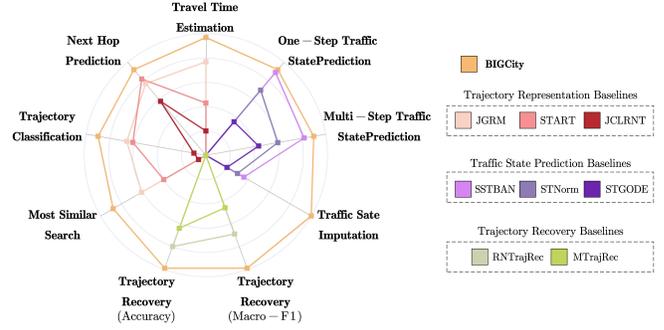}
     \caption{Performance Radarchart of \name on Various Tasks.}
     \label{fig:intro}
\end{figure}

In the literature, current ST data analysis methods are often narrowly tailored to specific types of data, focusing either on trajectories or traffic states. Most studies treat trajectories and traffic states as entirely distinct data modalities, designing task-specific models for one modality ({\bf \em sole task in sole data modality (STSD)}). A model designed for trajectory next-hop prediction cannot be applied to tasks like trajectory-user linkage, let alone traffic state prediction. While some ST data pre-training models, such as trajectory or time-series representation learning models, can generate universal representation vectors for multiple downstream tasks~\cite{START, liu2023itransformer}, they remain limited to handling a single data modality, either trajectories or traffic states ({\bf \em multiple tasks in sole data modality (MTSD)}).

However, individual-level mobility behaviors and population-level traffic states are intrinsically interconnected. Macro-level traffic states are aggregated from micro-level individual trajectories, while population traffic states influence individual mobility patterns. Many ITS and LBS applications require analyzing trajectory and traffic state data jointly. For example, car-hailing platforms need to predict the location of a taxi while considering traffic speed predictions. Treating trajectories and traffic states as separate domains in such scenarios can lead to suboptimal model performance.

In recent years, significant efforts have been made to develop universal foundation models for ST data. Models such as UniST~\cite{yuan2024unist}, UrbanGPT~\cite{li2024urbangpt}, and OpenCity~\cite{li2024opencity} focus on unified frameworks for traffic state analysis across cities, while UniTraj~\cite{zhu2024unitraj} and TrajCogn~\cite{zhou2024plm4trajcognizingmovementpatterns} aim to handle diverse trajectory analysis tasks using a single model. CityFM~\cite{balsebre2023cityfm} generates universal representations for static geographical units.
Despite their success in supporting multiple tasks~\cite{zhu2024unitraj}, cross-dataset knowledge transfer~\cite{yuan2024unist}, and few-/zero-shot data analysis~\cite{li2024opencity}, these models remain limited to {\em MTSD}. A truly versatile model capable of handling diverse ST data analysis tasks across both trajectories and traffic states ({\bf \em multiple tasks in multiple data modalities (MTMD)}) is still lacking.

Outside the field of ST data analysis, versatile and universal foundation models capable of processing heterogeneous data with a single framework have become widespread in domains such as text processing and image processing~\cite{zhao2023survey}. Notable examples include the success of large language models (LLMs) like~\cite{brown2020language, radford2019language} in natural language processing and visual-language multimodal data analysis~\cite{wang2023cogvlm}.
Despite the achievements of versatile models in these fields, developing a MTMD model for ST data remains constrained by several unique challenges, as outlined below.

\textbf{(1) Challenges in unified representations of ST data:} In NLP, all text data can be uniformly represented as sequences of characters (unified units), and in CV, images are uniformly represented as matrices or tensors of pixels (unified units). However, in ST data analysis, trajectories are modeled as sequences of geographical units (\eg road segments or POIs), while traffic states are represented as graphs with dynamic signals (\eg traffic speed over road networks). The differing basic units make developing a unified representation framework for the two data modalities a significant challenge.

\textbf{(2) Challenges in unifying heterogeneous ST analysis tasks:}
In NLP, diverse tasks can be unified as next-word generation, enabling a single model to handle various tasks efficiently. However, in ST data analysis, tasks are highly heterogeneous, with even identical inputs requiring different outputs. For instance, given the same trajectory inputs, travel time estimation outputs continuous timestamps, while user-trajectory linkage outputs discrete user IDs. Unifying such heterogeneous tasks within a single data processing and model training framework remains a significant challenge.

To address these challenges, we propose a {\underline{B}}i-modality un{\underline{I}}fied {\underline{G}}eneral model for ST data analysis in road network-based City scenarios (\name). \name focuses on road networks as the foundational scenario for ST data analysis. To tackle the challenge of unified data representation, \name introduces {\em ST-units} (see Sec.~\ref{sec:stunit}), enabling the expression of ST data from both trajectories and traffic state series in a unified format, \ie a sequence of ST-units. These ST-units are further embedded into token sequences via a neural network-based {\em Spatiotemporal Tokenizer} (see Sec.~\ref{sec:ST_tokenizer}).
To address the challenge of task heterogeneity, we propose a {\em Universal ST Model with Task-oriented Prompt (\submodel)} (see Sec.~\ref{model_structure}), which employs novel {\em Task-oriented Prompts} to guide an {\em LLM-based Tunable Model} for executing diverse ST analysis tasks. Additionally, we design a two-stage {\em training strategy} (see Sec.~\ref{sec:training}), combining self-supervised masked reconstruction and task-oriented prompt tuning, to train the model on heterogeneous ST data for multiple tasks.
Extensive experiments on three real-world datasets involving both trajectory and traffic state data demonstrate that \name outperforms 18 baselines across 8 tasks, highlighting its superior performance and versatility.

Here, we summarize our key contributions:
\begin{itemize}[leftmargin=*]
\item To the best of our knowledge, \name is the first model capable of handling multiple types of data analysis tasks for both trajectory and traffic state data, making it the first {\em MTMD} ST data analysis model.
\item We propose a unified ST data representation method, \ie ST-units and ST tokens, along with a novel universal model with task-oriented prompts to unify heterogeneous ST analysis tasks. These methods address the challenges of constructing unified representations for distinct ST data modalities and adapting to heterogeneous ST analysis tasks.
\item Our model achieves state-of-the-art performance on 3 real-world datasets, surpassing 18 baselines across 8 tasks. To our knowledge, \name is the first model to achieve SOTA performance over such a wide range of baselines and tasks.
\end{itemize}

%% file: content/6_related_work.tex
\section{Related Works}

\subsection{Spatiotemporal Data Analysis Models (STSD Models)} 
STSD models are trained to capture specific temporal and spatial dependencies, addressing a specific task such as traffic state prediction~\cite{jiang2023pdformer, wang2022traffic, ji2022stden,  wang2017coupling, wang2016traffic}, trajectory prediction~\cite{jiang2023continuous, wang2021personalized, wu2019learning, wang2019deep}, trajectory classification~\cite{wang2019empowering, wang2018cd}.
As a result, the model architectures vary with task type. For temporal dependency modeling, current models employ recurrent neural networks~\cite{li2017diffusion,bai2020adaptive}, temporal convolutional networks~\cite{wu2019graph,ji2020interpretable}, or attention mechanism~\cite{yao2019revisiting}. For spatial dependency modeling, they utilize graph neural networks~\cite{guo2021learning, yu2017spatio, wu2020connecting} or attention mechanism over graphs~\cite{ji2022precision}. Beyond the above models, physics-guided deep learning approaches have recently emerged, providing deeper theoretical insights into spatiotemporal data analysis (STSD)~\cite{hettige2024airphynet, ji2023self, ji2023multi, liufull, ji2020interpretable}. These models provide stronger interpretable ability, addressing limitations of deep models.

{\bf \em Summary:} Despite their success, most of the existing methods are specifically designed for a single data modality and specific tasks, \ie STSM. They requires separate models to handle trajectory and traffic state data. In contrast, \name can handle trajectories and traffic states within a single model.

\subsection{Spatiotemporal Representation learning (MTSD Models)}~\label{sec:STRL}
Self-supervised pre-training representation learning is crucial for multi-task ST data analysis. Existing trajectory models~\cite{chen2021robust, jiang2023self, lin2023pre, liu2022cstrm, mao2022jointly, yang2021unsupervised, yang2022weakly, zhu2022self, yang2023lightpath} employ sequential models with self-supervised tasks, while traffic state models~\cite{jiang2023unified, guo2019attention, han2021dynamic, jiang2023pdformer, li2017diffusion, wu2020connecting, wu2019graph, zheng2020gman,  ji2023spatio, wu2020learning} leverage graph neural networks (GNNs)~\cite{jin2023spatio} to capture spatiotemporal dependencies for predicting future traffic states. Recent models have focused on integrating more comprehensive ST features. For example, trajectory models like TremBR~\cite{Trembr} and START~\cite{jiang2023self} account for temporal periodicity, while traffic state models such as T-wave~\cite{9679147}, TrajNet~\cite{10.1145/3447548.3467236}, and TrGNN~\cite{Li_Tong_Li_Jin_Huang_Hua_2021} capture multi-hop spatial dependencies along trajectories. Beyond these models, VecCity~\cite{zhang2024veccity} proposed a comprehensive ST models library, where each map entity is represented as a vector. Furthermore, recent works~\cite{balsebre2023cityfm, yan2024urbanclip} have introduced text representations using large language models.
%\vspace{-0.1cm}

{\bf \em Summary:} While existing models often overlap in the trajectory and traffic state domains, such as incorporating traffic state into trajectory representations, they are typically focused on a single data modality, \ie {\em MTSD} models. Furthermore, these models require fine-tuning for different downstream tasks, resulting in ``semi-multi-task'' functionality. While \name handles various tasks without task-specific fine-tuning.

\subsection{Universal Spatiotemporal Models (ST Foundation Models)}
%\vspace{-0.3cm}

The target of this kind of model is to design a universal foundation model for ST data analysis, similar to LLMs for NLP. There are two main research directions for spatiotemporal universal models: 1) {\em Cross-Dataset Universal Models}: These models aim to generalize across different datasets (\eg datasets from various cities), enabling rapid adaptation to new datasets through few-shot or zero-shot learning~\cite{feng2024citygpt, feng2024citybench, li2024urbangpt, xu2023urban, yuan2024unist, DBLP:journals/corr/abs-1909-12756, jin2023time, zhou2023one, shao2022pre, li2024opencity, li2024flashst} for traffic states or trajectory analysis. For example, approaches~\cite{liu2024spatial, ren2024tpllm} treat LLMs as spatio-temporal encoders, training them for traffic prediction over cross-data sets. UniST~\cite{yuan2024unist} introduced spatiotemporal prompt learning, using statistical features of the dataset as prompts for the prediction of cross-dataset traffic. 2) {\em Universal Models for Heterogeneous Tasks:} These models focus on multi-task capabilities, with one strategy involving training from scratch on large datasets~\cite{zhang2023regions} to adapting heterogeneous with the same data modalities. Recent works~\cite{villarreal2023can, da2023llm, zheng2023chatgpt, wang2023would, xue2022leveraging, zhang2023trafficgpt, zhou2024plm4trajcognizingmovementpatterns} exploit LLMs' multi-task capabilities by converting trajectory data into textual format. Other works leverage LLMs' linguistic abilities to improve model generalization via in-context learning~\cite{li2024urbangpt} and provide interpretable outputs~\cite{guo2024towards}. Some efforts explore LLMs as spatiotemporal agents, bridging LLMs with tools to perform traffic-related queries and reasoning based on human instructions~\cite{da2024open}.
%\vspace{-0.2cm}

{\bf \em Summary:} Many of these universal ST models leverage LLM technologies for task versatility and cross-dataset knowledge transfer. These models effectively address the ``semi-multi-task'' limitation of ST representation learning by eliminating the need for fine-tuning in downstream applications. However, most of these models are limited to handling either trajectory or traffic state data, but not both. In contrast, \name not only addresses the ``semi-multi-task'' issue by offering task versatility but also handles both trajectory and traffic state data, achieving true data modality versatility.

%% file: content/wang_version/2_pre_wang.tex
% \begin{figure}
%     \centering
%     \includegraphics[width=1.0\linewidth]{figure/2_preliminary/STUnit.pdf}
%     \caption{A unified expression for spatiotemporal data}
%     \label{STUnit}
%     \vspace{-0.2cm}
% \end{figure}

\section{Preliminaries}
\label{sec:pre}

Our model is designed for road network-based urban traffic scenarios, where the city is represented as a road network and ST data are generated from individuals' movements. These scenarios are common in many widely used ST datasets. This section outlines the {\em basic spatial and temporal components} of such scenarios, defining the two types of {\em dynamic spatiotemporal data}: trajectories and traffic states.

\subsection{Basic Spatial and Temporal Elements}
\label{sec:elements}

The basic spatial elements in the scenario are road segments.

\begin{mydef}[Road Segment]~\label{def:segment}
    Consider a city map with $I$ road segments, denoted as $r_i$ for the $i$-th segment. The set of segments is {\small $\mathcal{R} = \{r_1, \cdots, r_i, \cdots, r_I\}$}. Each segment $r_i$ is associated with a static feature vector {\small $\bm{e}_i^{(s)} \in \mathbb{R}^{D_r}$}, describing attributes such as road ID, type, length, lane count, in-degree, out-degree, speed limit, and other relevant characteristics.
\end{mydef}

All road segments collectively form a road network.

\begin{mydef}[Road Network]~\label{def:network}
    A road network is a directed graph denoted as {\small $\mathcal{G} = \{\mathcal{R}, \mathcal{A}, \bm{E}^{(s)}\}$}, where $\mathcal{R}$ is the set of vertices corresponding to road segments. {\small $\mathcal{A} \in \mathbb{R}^{|\mathcal{R}|\times|\mathcal{R}|}$} is the binary adjacency matrix indicating connectivity between road segments. {\small $\bm{E}^{(s)} = \big(\bm{e}_1^{(s)}, \cdots, \bm{e}_n^{(s)}, \cdots, \bm{e}_N^{(s)}\big)$} represents the static feature vectors for all road segments.
\end{mydef}

The basic temporal elements are conceptualized in two forms: discrete time slice and continuous timestamp.

\begin{mydef}[Time Slice]~\label{def:timeslice}
    A time slice is a fixed-length interval partitioning the timeline, indexed as $\{1, \cdots, t, \cdots, T\}$. For the $t$-th time slice, a feature vector $\bm{\iota}_{t} \in \mathbb{R}^{D_t}$ is defined to describe its attributes, such as the slice's start time, its index within a day, the day index within a week, and so on.
\end{mydef}

\begin{mydef}[Timestamp]~\label{def:timestamp}
    A timestamp represents an instantaneous UTC time, denoted as $\tau$. For a given timestamp $\tau$, a feature vector $\bm{\iota}_{\tau} \in \mathbb{R}^{D_{\tau}}$ describes its attributes, including its absolute timeand the features of the time slice it belongs to.
\end{mydef}

In \name, both discrete and continuous temporal elements coexist. For a time slice $t$, its start time is represented by the timestamp $\tau_t$, and for a timestamp $\tau$, the corresponding time slice is denoted as $t_\tau$.

\subsection{Dynamic Spatiotemporal Data}

Based on the above elements, dynamic ST data can be categorized into two types: individual-level trajectories and population-level traffic states. Trajectories capture individual mobility behaviors, defined as follows:

\begin{mydef}[Trajectory]~\label{def:traj}
    A trajectory is a time-ordered sequence of road segments with associated timestamps, defined as {\small ${tr} = \big( (r_{tr_1}, \tau_{tr_1}),$ $\dots,$ $(r_{tr_l}, \tau_{tr_l}),$ $\dots,$ $(r_{tr_L}, \tau_{tr_L}) \big)$}, where {\small $(r_{tr_l}, \tau_{tr_l})$} is the $l$-th sample in the trajectory. Here, {\small $r_{tr_l} \in \mathcal{R}$} is the road segment, and {\small $\tau_{tr_l}$} is the corresponding timestamp. The trajectory can also be expressed as {\small ${tr} = \big( (\bm{e}^{(s)}_{tr_1}, \tau_{tr_1}),$ $\dots,$ $(\bm{e}^{(s)}_{tr_l}, \tau_{tr_l}),$ $\dots,$ $(\bm{e}^{(s)}_{tr_L}, \tau_{tr_L}) \big)$}, where {\small $\bm{e}^{(s)}_{tr_l}$} denotes the static feature of road segment {\small $r_{tr_l}$}.
\end{mydef}

% A raw trajectory consists of a sequence of geographic coordinates paired with timestamps. To transform these into a road-segment-based trajectory, we use map-matching algorithms~\cite{yang2018fast} to project the coordinates onto road segments.

\begin{mydef}[Traffic State]~\label{def:ts}
    For a given time slice $t$, the traffic state of the road segment $r_i$ is represented as a vector {\small $\bm{e}^{(d)}_{i,t} \in \mathbb{R}^{D_d}$}, contains the {\em dynamic characteristics} of $r_i$, such as average speed, traffic entry and exit. $D_d$ is the number of traffic state channels. The traffic state series for $r_i$ is {\small ${ts}_i = \big(\bm{e}^{(d)}_{i,1},$ $\dots,$ $\bm{e}^{(d)}_{i,t},$ $\dots,$ $\bm{e}^{(d)}_{i, T}\big)$}. Using the start time of each time slice as its timestamp, we have {\small ${ts}_i = \big( (\bm{e}^{(d)}_{i,1}, \tau_1),$ $\dots,$ $(\bm{e}^{(d)}_{i,t}, \tau_t),$ $\dots,$ $(\bm{e}^{(d)}_{i,T}, \tau_T) \big)$}, where $\tau_t$ is the timestamp of time slice $t$.
\end{mydef}

According to Def.~\ref{def:traj} and~\ref{def:ts}, both trajectories and traffic states are sequences of the basic spatial and temporal elements.

\begin{figure}
    \centering
    \includegraphics[width=1.0\linewidth]{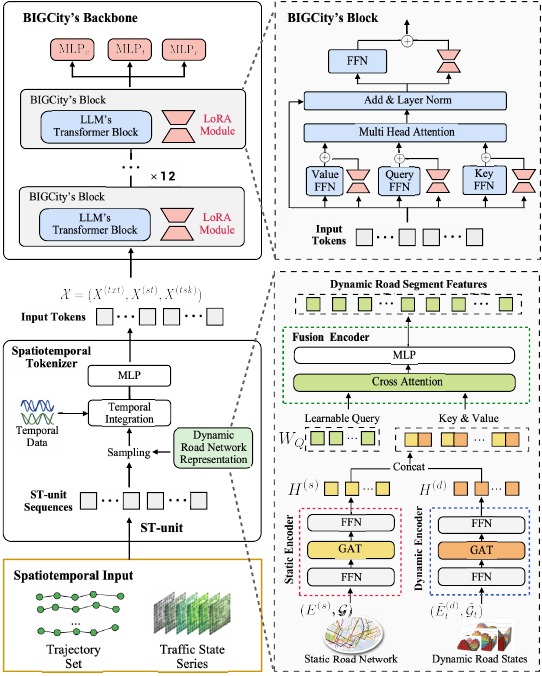}
    \caption{The model architecture of \name.}\label{fig:model_architecture}
\end{figure}

\subsection{Motivation of \name}~\label{sec:motivation}
Trajectories and traffic states represent human mobility patterns at different levels: individual-level and population-level, respectively. Although typically treated as distinct, heterogeneous data modalities, these two types of data are inherently interconnected. According to Def.~\ref{def:traj} and Def.~\ref{def:ts}, both trajectories and traffic states share similar structural formats.
\begin{itemize}[leftmargin=*]
\item {\em For a trajectory}, a sample corresponds to the \underline{{\em road segment}} where an individual is located at a given \underline{{\em sampling time}}, and the location is associated with a corresponding \underline{{\em traffic state}}.
\item {\em For a traffic state series}, a sample represents the dynamic \underline{{\em traffic state}} of a \underline{{\em road segment}} at a given \underline{{\em sampling time}}.
\end{itemize}
Thus, the triple $(\mathrm{segment}, \mathrm{traffic~state}, \mathrm{sampling~time})$, \ie ``a road segment with its traffic state sampled at a specific time'', can be seen as the basic unit of ST data, analogous to words in NLP or pixels in image data. This insight suggests a viable opportunity to develop a versatile {\em MTMD} model that simultaneously handles both trajectories and traffic states.

Based on this idea, we propose \name as to achieve {\em MTMD} ST modeling. Figure~\ref{fig:model_architecture} illustrates the framework, which consists of two core components: $i$) a {\em Unified ST Tokenizer} to address heterogeneous data representation (Sec.~\ref{sec:unifieddata}), and $ii$) a {\em Versatile ST Model with Task-oriented Prompt} to handle distinct ST tasks (Sec.~\ref{model_structure}).

%% file: content/wang_version/3_2_tokenizer.tex
\section{Unified Representations for ST Data}~\label{sec:unifieddata}
In this section, we address {\em Challenge 1}: the unification of spatiotemporal data representations for MTMD model design. We first define {\em spatiotemporal units} (ST-units) as a unified representation for both trajectories and traffic states. Then, we introduce the {\em Spatiotemporal Tokenizer} to encode these ST-units into input tokens (ST tokens) for \name.

%To address {\em Challenge 1}, we proposed a unified representation approach for multi-level ST data. Specifically, we first define {\em basic spatiotemporal units} (ST-units), representing both trajectories and traffic states in a unified format, \ie sequences of ST-units. Then, we propose the {\em Spatiotemporal Tokenizer} to encode these ST-units into input tokens (ST tokens) for \name backbone.

\subsection{Basic Spatiotemporal Units}~\label{sec:stunit}
Building on the motivation in Sec.~\ref{sec:motivation}, we define the triple $(\mathrm{segment}, \mathrm{traffic~state}, \mathrm{sampling~time})$ as the basic spatiotemporal unit (ST-unit) for both trajectory and traffic state data. Formally, for a road segment $r_i$ and a timestamp $\tau$, an ST-unit is expressed as:
\begin{equation}\label{eq:STU}\small
  \bm{U}_{i,\tau} = \left( \bm{e}^{(s)}_{i}, \bm{e}^{(d)}_{i,t_\tau}, \bm{\iota}_\tau \right),
\end{equation}
where $\bm{e}^{(s)}_{i}$ represents the static features of segment $r_i$, $\bm{\iota}_\tau$ is the timestamp feature for $\tau$, and $\bm{e}^{(d)}_{i,t_\tau}$ represents the dynamic traffic state of segment $r_i$ at the time slice containing $\tau$.

Using the ST-unit in Eq.~\eqref{eq:STU}, we redefine traffic states and trajectories in a unified format as sequences of ST-units. %Specifically, we define the traffic state series as follows:

\begin{mydef}[ST-unit-based Traffic State]~\label{def:traffic_series_unit}
For a road segment $r_i$, its traffic state series is redefined as:
\begin{equation}\label{eq:traffic_series}\small
  \mathbf{\mathcal{U}}_{i} = \Big( \bm{U}_{i,\tau_{1}}, \cdots, \bm{U}_{i,\tau_{t}}, \cdots, \bm{U}_{i,\tau_{T}} \Big),
\end{equation}
where {\small $\bm{U}_{i,\tau_{t}} = \big(\bm{e}^{(s)}_{i}, \bm{e}^{(d)}_{i,t}, \bm{\iota}_{\tau_{t}}\big)$}. Here, {\small $\bm{\iota}_{\tau_{t}}$} represents the features of the timestamp $\tau_{t}$, and $\tau_{t}$ is the start time of time slice $t$.
\end{mydef}

\begin{mydef}[ST-unit-based Trajectory]~\label{def:trajectory_unit}
Given a trajectory $tr$ of length $L$, its ST-unit-based representation is defined as:
\begin{equation}\label{eq:trajectory}\small
  \mathbf{\mathcal{U}}_{{tr}} = \Big( \bm{U}_{{tr}_1}, \cdots, \bm{U}_{{tr}_l}, \cdots, \bm{U}_{{tr}_L} \Big),
\end{equation}
where {\small $\bm{U}_{{tr}_l} = \big(\bm{e}^{(s)}_{tr_l}, \bm{e}^{(d)}_{tr_l, t_{tr_l}}, \bm{\iota}_{\tau_{tr_l}}\big)$} denotes the ST-unit for the $l$-th sample. Here, {\small $\bm{e}^{(s)}_{tr_l}$} represents the static features of road segment $tr_l$, {\small $\bm{e}^{(d)}_{tr_l,t_{tr_l}}$} the dynamic features at time slice $t_{tr_l}$, and {\small $\bm{\iota}_{\tau_{tr_l}}$} the timestamp feature of $\tau_{tr_l}$.
\end{mydef}

In certain datasets, road segments may lack dynamic features. In such cases, we set {\small $\bm{e}^{(d)}_{tr_l, \tau_{tr_l}} = \mathrm{NULL}$}.

\paratitle{\em Remark:} In Eq.~\eqref{eq:trajectory} and Eq.~\eqref{eq:traffic_series}, both individual-level trajectories and population-level traffic states, despite their heterogeneous modalities, are represented in a unified format -- sequences of ST-units. This unification resolves the challenge of heterogeneous data representation, enabling the design of a model capable of analyzing multiple ST data modalities.

\subsection{Spatiotemporal Tokenizer}~\label{sec:ST_tokenizer}
This subsection presents the spatiotemporal (ST) tokenizer, which converts ST-units into token vectors (ST-tokens). Specifically, it first generates dynamic road network representations as ST feature library, and then samples specific features for input data according to ST-units. The tokenizer comprises four modules: a static feature encoder, a dynamic feature encoder, a fusion encoder, and a temporal integration module.

\paratitle{Static Feature Encoder.} This module encodes the static features {\small $\bm{e}^{(s)}_{i}$} of an ST-unit into a representation vector. To capture the spatial and topological relationships among road segments, the encoder employs a graph attention network (GAT)~\cite{velivckovic2017graph}, generating segment representations based on the road network {\small $\mathcal{G} = \{\mathcal{R}, \mathcal{A}, \bm{E}^{(s)}\}$} defined in Def.~\ref{def:network}. Specifically, for the static feature matrix {\small $\bm{E}^{(s)} = \big( \bm{e}^{(s)}_1, \dots, \bm{e}^{(s)}_N\big)$}, the encoder outputs a representation matrix {\small $\bm{H}^{(s)}$} as follows:
\begin{equation}\label{eq:GAT1_s}\small
  \bm{H}^{(s)} = \mathrm{FFN}\left( \mathrm{GAT}_s \big(\bm{E}^{(s)}, \mathcal{G}\big)\right),
\end{equation}
where {\small $\mathrm{GAT}_s(\cdot, \cdot)$} is the GAT model and {\small $\mathrm{FFN}(\cdot)$} a feed-forward network for dimensional transformation. The result is {\small $\bm{H}^{(s)} = \big(\bm{h}^{(s)}_1, \dots, \bm{h}^{(s)}_I\big)$}, with {\small $\bm{h}^{(s)}_i \in \mathbb{R}^{D_h}$} as the {\em static representation} for road segment $r_i$.

\paratitle{Dynamic Feature Encoder.} This module encodes the dynamic features {\small $\bm{e}^{(d)}_{i,\tau}$} of an ST-unit into a representation vector. To capture temporal dependencies, historical features are incorporated from a time window of length $T'$. For time slice $t$, the window is defined as {\small $W = (t-T', \cdots, t-1, t)$}, and the concatenated historical features for segment $r_i$ are {\small $  \tilde{\bm{e}}^{(d)}_{i,t} = \left(\bm{e}^{(d)}_{i,t-T'} \big\| \cdots \big\| \bm{e}^{(d)}_{i,t}\right)$},
%\begin{equation}\small
%  \tilde{\bm{e}}^{(d)}_{i,t} = \left(\bm{e}^{(d)}_{i,t-T'} \big\| \cdots \big\| \bm{e}^{(d)}_{i,t}\right),
%\end{equation}
where {\small $\|$} denotes concatenation. The integrated historical dynamic feature matrix is {\small $\tilde{\bm{E}}^{(d)}_{t} = \big(\tilde{\bm{e}}^{(d)}_{1,t}, \cdots, \tilde{\bm{e}}^{(d)}_{I,t}\big)$}. By replacing the static feature matrix in the road network with {\small $\tilde{\bm{E}}^{(d)}_{t}$}, a dynamic road network {\small $\tilde{\mathcal{G}}_t = \{\mathcal{R}, \mathcal{A}, \tilde{\bm{E}}^{(d)}_{t}\}$} is constructed. The dynamic feature encoder uses a GAT to encode $\tilde{\mathcal{G}}_t$ as:
\begin{equation}\small
  \bm{H}^{(d)}_{t} = \mathrm{FFN} \Big( \mathrm{GAT}_d \big(\tilde{\bm{E}}^{(d)}_{t}, \tilde{\mathcal{G}}_{t}\big) \Big),
\end{equation}
where the result {\small $\bm{H}^{(d)}_t = \big(\bm{h}^{(d)}_{1,t}, \cdots, \bm{h}^{(d)}_{I,t}\big)$} represents the {\em dynamic representation} of segment $r_i$ at time slice $t$.

\paratitle{Fusion Encoder.} This module fuses static and dynamic representations of road segments to generate comprehensive spatial representations. The concatenated representation for segment $r_i$ at time slice $t$ is {\small $\bm{h}_{i,t} = \big(\bm{h}^{(s)}_{i}\| \bm{h}^{(d)}_{i,t}\big)$}. A cross-attention mechanism is employed to capture long-range dependencies among these representations. Specifically, for segments $r_i$ and $r_j$, their relationship is computed as:
\begin{equation}\small
  \alpha_{ij} = \Big({\bm{q}_{i}^\top \bm{h}_{j,t}}\Big)\Big/{\sqrt{2 D_h}},
\end{equation}
where {\small $\bm{W}_Q \in \mathbb{R}^{ I \times D_h}$} is a learnable query matrix, and $q_i$ is the $i$-th vector The fused {\em spatial representation} for $r_i$ at time slice $t$ is then:
\begin{equation}\small
  \bm{s}_{i,t} = \sum_{j=1}^{I} \mathrm{ATT}_{ij} \cdot \bm{h}_{j,t},\;\;\; \mathrm{where} \;\;\; \mathrm{ATT}_{ij} = {\alpha_{ij}}\Big/{\sum_{j=1}^{I} \alpha_{ij}}.
\end{equation}
Unlike the GAT used in the static and dynamic encoders, which only capture correlations between directly connected segments, the cross-attention mechanism enables long-range dependencies across all segments.

\paratitle{Temporal Integration \& ST Tokens.} This module integrates the timestamp and its features with the spatial representation to generate an ST token. Specifically, for a ST-unit {\small $\bm{U}_{i,\tau} = \big( \bm{e}^{(s)}_{i}, \bm{e}^{(d)}_{i,t_\tau}, \bm{\iota}_\tau \big)$}, the static and dynamic features {\small $\bm{e}^{(s)}_{i}$} and {\small $\bm{e}^{(d)}_{i,t_\tau}$} are encoded into a spatial representation {\small $\bm{s}_{i,t_\tau}$}. An MLP then combines {\small $\bm{s}_{i,t_\tau}$} with the timestamp feature {\small $\bm{\iota}_\tau$} and the time interval {\small $\delta_\tau$} between adjacent ST-units to generate the ST token:
\begin{equation}\label{eq:st_tokens}\small
  \bm{x}_{i,\tau} = \mathrm{MLP}\Big( \; \bm{s}_{i,t_\tau} \big\| \bm{\iota}_\tau \big\| \delta_\tau \Big),
\end{equation}
where {\small $\delta_\tau$} is the time interval between consecutive ST-units in a sequence. For a sequence of ST-units {\small $(\bm{U}_{1}, \cdots, \bm{U}_{l}, \cdots, \bm{U}_{L})$} corresponding to a trajectory or traffic state series, the interval is {\small $\delta_{\tau_l} = \tau_{l} - \tau_{l-1}$}. Including {\small $\delta_{\tau_l}$} helps the model handle non-uniformly spaced ST-unit sequences, which is crucial for real-world trajectory data with irregular sampling intervals. The resulting vector {\small $\bm{x}_{i,\tau}$} is the {\em ST Token} for ST-unit {\small $\bm{U}_{i,\tau}$}.

\paratitle{\em Remark:} The ST tokenizer converts a sequence of ST-units, representing either a trajectory or a traffic state series, into an ST-token sequence. Specifically, it transforms the trajectory {\small $\mathcal{U}_{\mathrm{tr}}$} (Eq.~\eqref{eq:trajectory}) into {\small $\bm{X}_{tr} = \big(\bm{x}_{tr_1}, \cdots, \bm{x}_{tr_L}\big)$}, and the traffic state series {\small $\mathcal{U}_i$} (Eq.~\eqref{eq:traffic_series}) into {\small $\bm{X}_i = \big(\bm{x}_{i,\tau_1}, \cdots, \bm{x}_{i,\tau_T}\big)$}. These ST-token sequences serve as unified inputs for the \name model, allowing it to process heterogeneous ST data across modalities. Thus, both individual-level trajectories and population-level traffic states are represented in a unified form, enabling seamless processing by a single model.

\begin{figure*}[t]
\captionsetup{font=small}
    \centering
    \captionsetup[subfigure]{labelformat=simple}
    \begin{subfigure}[a]{0.23\textwidth}
        \includegraphics[width=\textwidth]{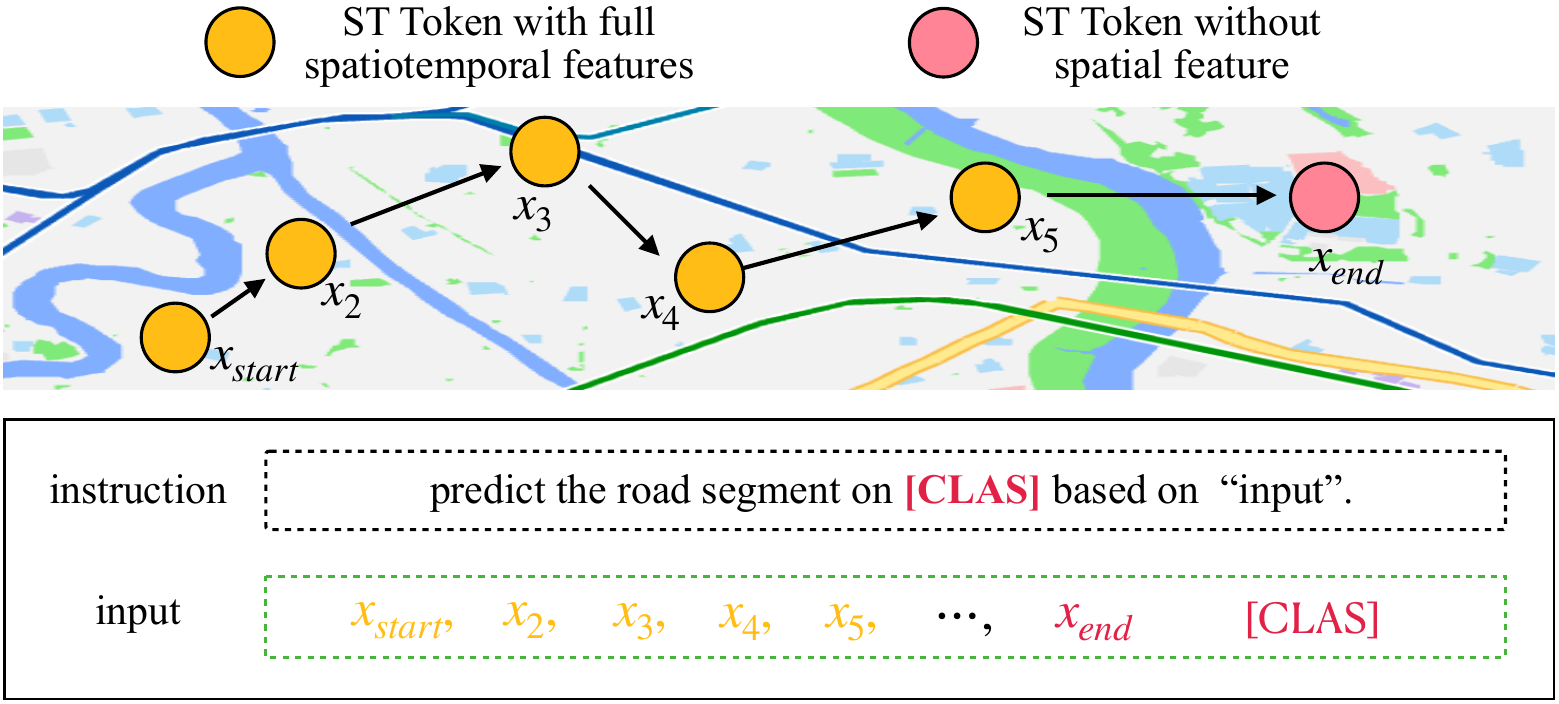}
        % \vspace{-0.3cm}
        \caption{ \scriptsize Template of Next Hop Prediction}
        \label{fig:next_hop_template}
    \end{subfigure}
    \hfill\hfill
    \captionsetup[subfigure]{labelformat=simple}
    \begin{subfigure}[a]{0.23\textwidth}
        \includegraphics[width=\textwidth]{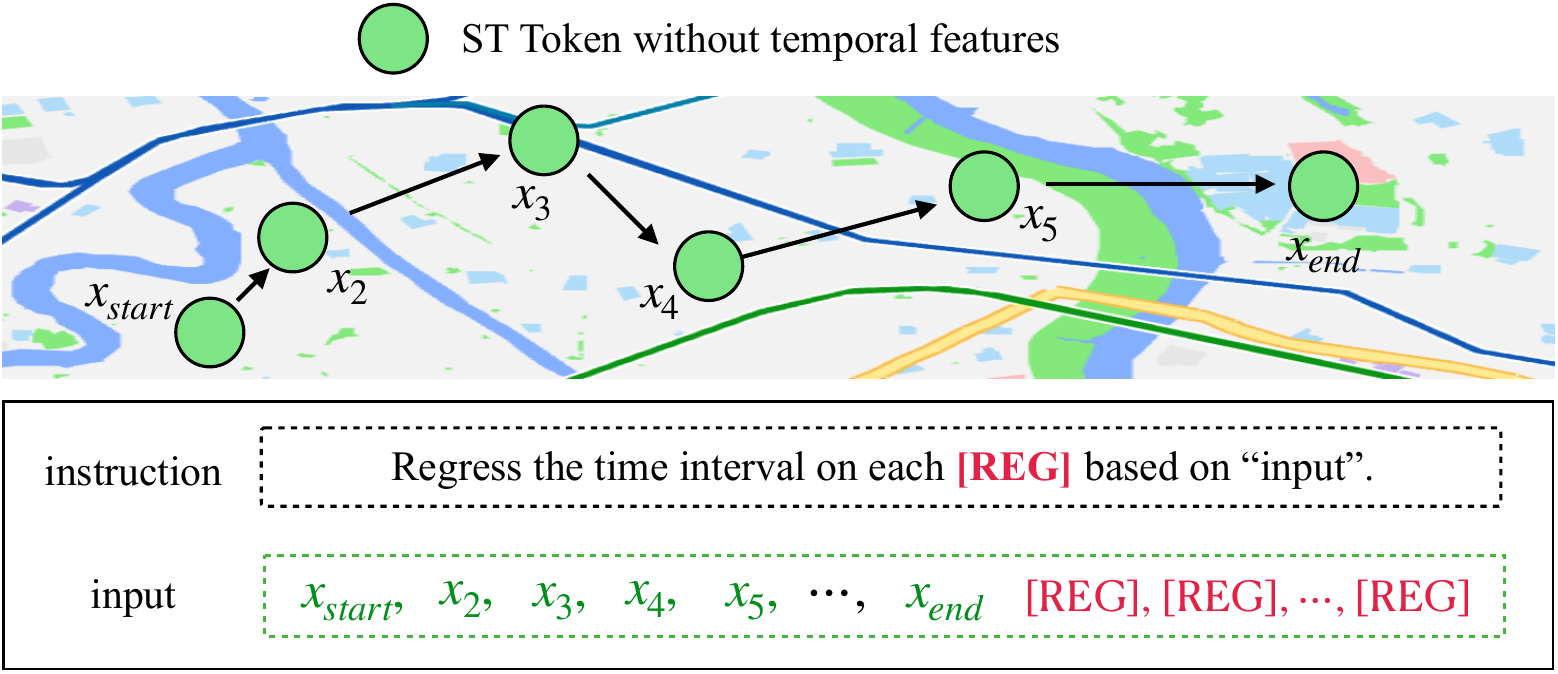}
        % \vspace{-0.3cm}
        \caption{ \scriptsize Template of TTE Task}
        \label{fig:TTE_template}
    \end{subfigure}
    \hfill\hfill
    \captionsetup[subfigure]{labelformat=simple}
    \begin{subfigure}[a]{0.205\textwidth}
        \includegraphics[width=\textwidth]{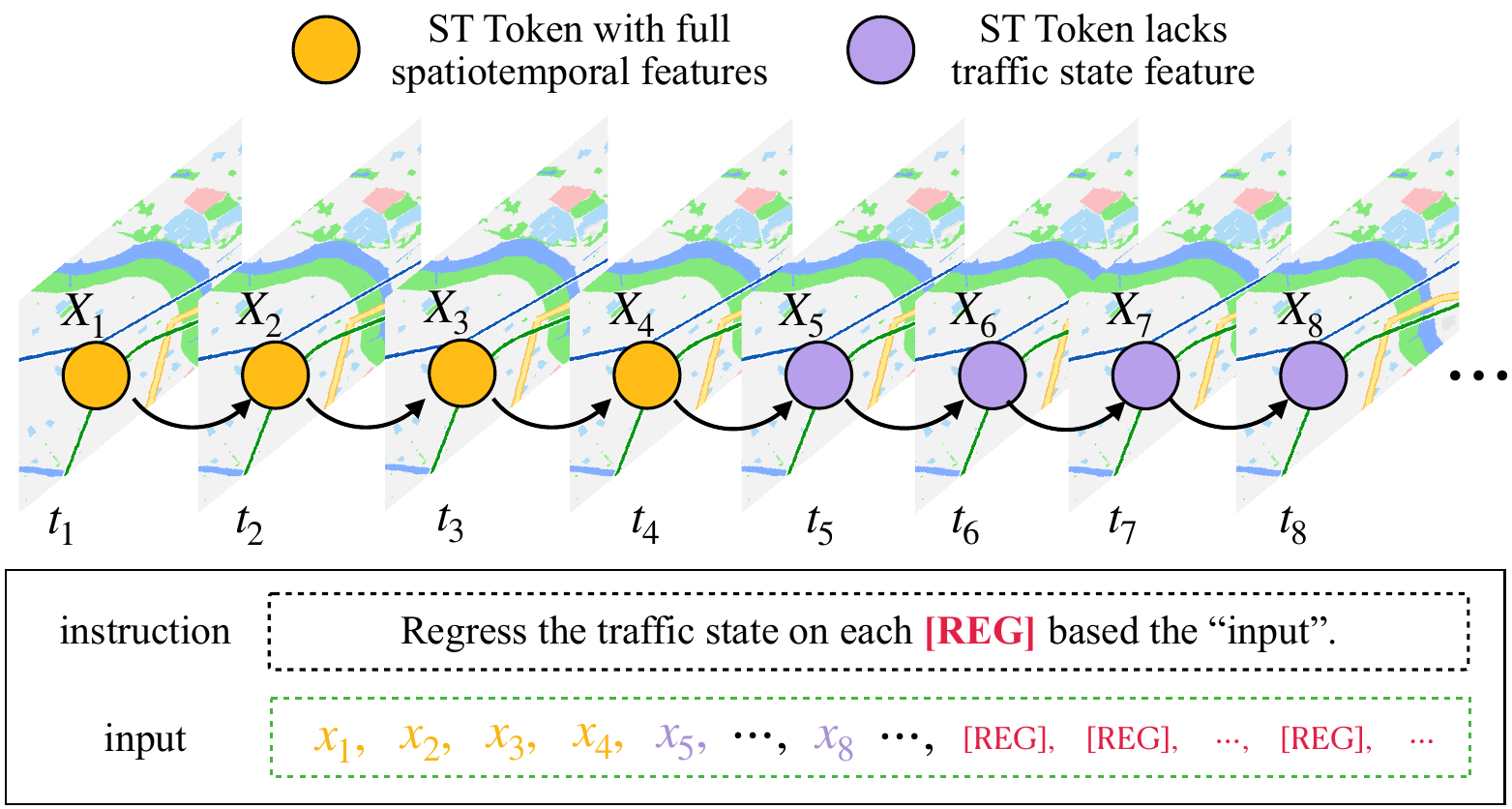}
        % \vspace{-0.3cm}
        \caption{ \scriptsize Template of Traffic State Prediction}
        \label{fig:traffic_state_template}
    \end{subfigure}
    \hfill\hfill
    \captionsetup[subfigure]{labelformat=simple}
    \begin{subfigure}[a]{0.23\textwidth}
        \includegraphics[width=\textwidth]{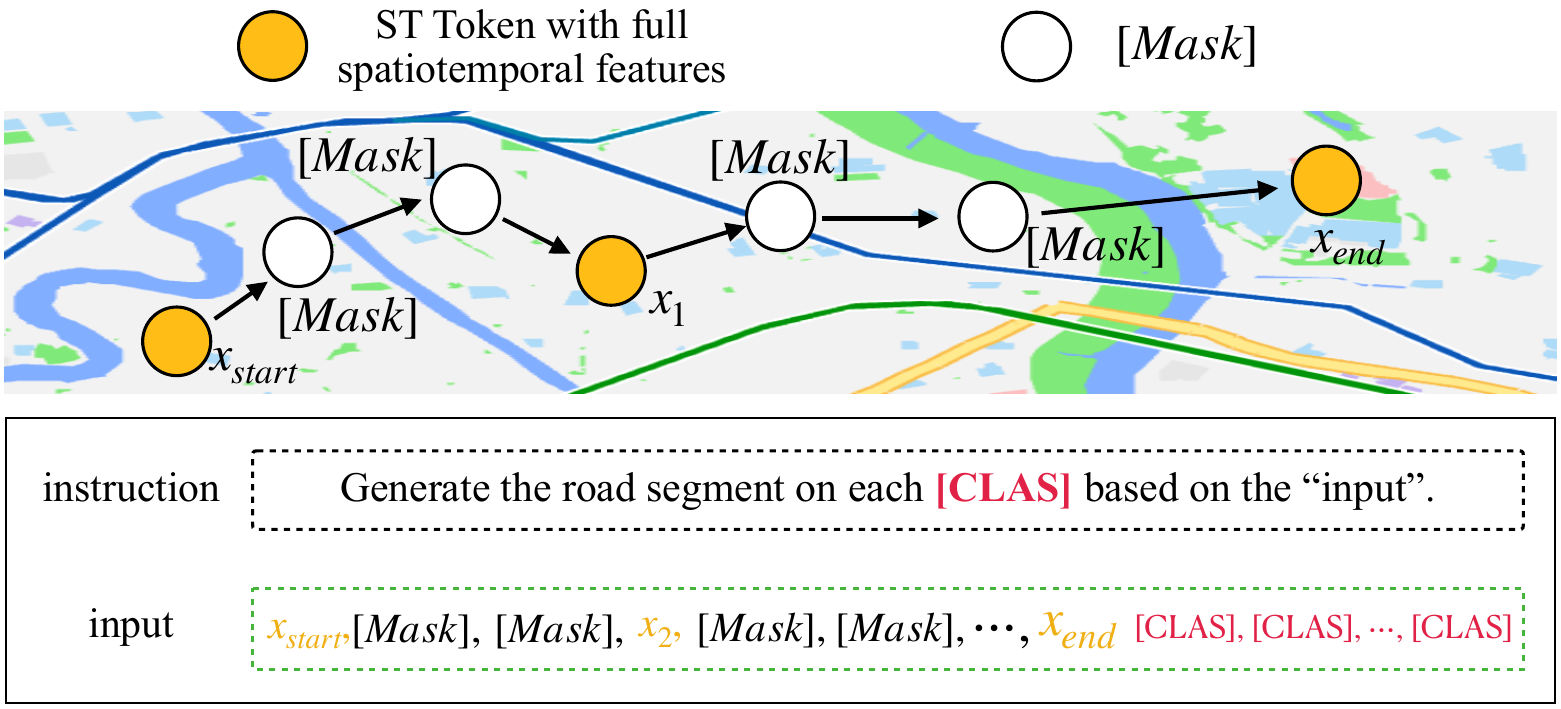}
        % \vspace{-0.3cm}
        \caption{ \scriptsize Template of Trajectory Recovery}
        \label{fig:trajectory_recovery}
    \end{subfigure}
    \hfill
    \caption{The templates for Task-oriented Prompts with different tasks.}
    % \vspace{-0.3cm}
    \label{fig:prompt_example}
\end{figure*}

%\begin{figure*}[t]\label{fig:prompt_example}
%     \centering
%     \subfigure[Template of Next Hop]{\label{fig:next_hop_template}
%     \includegraphics[width=0.3\linewidth]{figure/prompt_templates/NextHopPrediction.pdf}}
%     \subfigure[Template of Traffic State Prediction]{\label{fig:traffic_state_template}
%     \includegraphics[width=0.3\linewidth]{figure/prompt_templates/Traffic State.pdf}}
%     \subfigure[Template of Trajectory Recovery]{\label{fig:trajectory_recovery}
%     \includegraphics[width=0.3\linewidth]{figure/prompt_templates/Trajectory Recovery.pdf}}
%     %\vspace{-.3cm}
%     \caption{The prompt templates.}
%     \vspace{-.5cm}
%\end{figure*}

%% file: content/wang_version/3_3_backbone_model.tex
\section{Versatile Model with Task Oriented Prompt}
\label{model_structure}

This section introduces a method to address the challenge of adapting to diverse ST analysis tasks ({\em Challenge 2}) for designing an {\em MTMD} model. A key difficulty for this challenge lies in informing the model about the specific task it should perform. Even with identical data inputs, different tasks require distinct types of outputs. For instance, given the same trajectory inputs, the travel time estimation task outputs continuous arrival time predictions, while the user-trajectory linkage task outputs discrete user ID.

Drawing inspiration from the prompt mechanism in LLMs, we propose a {\em Versatile Model with Task-oriented Prompts (\submodel)} to address this difficulty. \submodel employs textual instructions as prompts to guide the model on the desired task and utilizes a tunable LLM with ST tokens as inputs to perform specific ST data analysis tasks. The model consists of three modules: {\em Task-oriented Prompts} (input module), {\em LLM-based Backbone Model} (data-processing module), and {\em General-task Heads} (output module).

%We introduce the three module the following. %\subsection{Tunable Backbone Large Model}~\label{sec:tunable_LLM_backbone}
%the versatility of

\subsection{Task-oriented Prompts}

The input to our model is a task-oriented prompt that combines the ST token sequences generated by the ST tokenizer in Sec.~\ref{sec:ST_tokenizer} with textual instructions specifying the type of task to be performed.

\paratitle{Prompt Contents.} The task-oriented prompt consists of three components:
\begin{itemize}[leftmargin=*]
\item {\em Textual Instructions.} This part of the prompt is a textual description of the data analysis task to be executed. For example, the instruction {\em ``Where is the next hop position of the input trajectory?''} informs the model to perform a next hop prediction task. We use the tokenizer of the backbone LLM (see Sec.~\ref{sec:backbone}) to converts the instructions into a sequence of {\em Text Tokens}, denoted as {\small $\bm{X}^{(\mathrm{txt})}$}.
\item {\em Input Data.} This part of the prompt provides the spatiotemporal data to be analyzed, such as a traffic state series or a trajectory, formatted as ST-unit sequences according to Eq.~\eqref{eq:traffic_series} or Eq.~\eqref{eq:trajectory}. The ST tokenizer, as described in Sec.~\ref{sec:ST_tokenizer}, converts these sequences into a series of {\em ST Tokens}, denoted as {\small $\bm{X}^{(\mathrm{st})}$}.
\item {\em Task Placeholders.} This part of the prompt provides a format guide for the task outputs. Two types of placeholders are used: the classification placeholder, denoted as {\small $[\mathrm{CLAS}]$}, and the regression placeholder, denoted as {\small $[\mathrm{REG}]$}. These placeholders represent the expected output structure of the ST data analysis task. We use two learnable token vectors, {\small $\bm{x}^{(\mathrm{clas})}$} and {\small $\bm{x}^{(\mathrm{reg})}$}, to corresponding to the two types of placeholders. The sequence of learnable token vectors is named as {\em Task Tokens} and denoted as {\small $\bm{X}^{(\mathrm{tsk})}$}.
\end{itemize}
The complete inputs to the \submodel model is a combined sequence consisting of the {text tokens} {\small $\bm{X}^{(\mathrm{txt})}$}, the {ST tokens} {\small $\bm{X}^{(\mathrm{st})}$}, and the {task tokens} {\small $\bm{X}^{(\mathrm{tsk})}$}, named as {\em input prompt tokens}, represented as:
\begin{equation}\label{}\small
    \bm{\mathcal{X}} = \left(\bm{X}^{(\mathrm{txt})}, \bm{X}^{(\mathrm{st})}, \bm{X}^{(\mathrm{tsk})}\right).
\end{equation}

\paratitle{Prompt Templates.} As illustrated in the examples in Fig.~\ref{fig:prompt_example}, we use a template to organize the {\em instructions}, {\em input data}, and {\em task placeholders} of task-oriented prompts into a unified structure for various tasks.

The first part of the template is the instruction. For each task, we start by using a language model, specifically ChatGPT, to understand the task's function. Then, the language model generates a set of candidate instructions describing the task. For example, for the travel time estimation task, candidate instructions might include: {\em ``Give me the estimated time of arrival for the input trajectory''} or {\em ``When will I walk to the ending position in this trajectory?''}. Finally, we evaluate these candidates and select the most effective instruction based on testing. This selected instruction is then used as a fixed component in the prompt template.

The second and third parts of the template are the input data and task placeholders. These components vary slightly depending on the type of task:
\begin{itemize}[leftmargin=*]
  \item {\em For classification tasks}, such as trajectory next hop prediction and user-trajectory linkage, the input data consists of a sequence of ST tokens corresponding to a trajectory to be classified. The task placeholder is a classification placeholder {\small $[\mathrm{CLAS}]$} (see Fig.~\ref{fig:next_hop_template}).

  \item {\em For regression tasks}, such as trajectory travel time estimation (TTE) and traffic state prediction, the input data is a sequence of ST tokens corresponding to a trajectory or traffic state series. The task placeholder is a sequence of regression placeholder{\small $[\mathrm{REG}]$} (see Fig.~\ref{fig:TTE_template} and Fig.~\ref{fig:traffic_state_template}). %Specifically, TTE and one-step prediction use a single placeholder, while multi-step prediction employs multiple placeholders.
  \item {\em For generation tasks}, such as trajectory recovery, the input consists of a sequence of ST tokens with {\small $[\mathrm{MASK}]$} inserted at the positions to be generated. In recovery, {\small $[\mathrm{MASK}]$} are placed between adjacent samples in a low-rate trajectory. The task placeholders are sequences of classification pairs, denoted as {\small $\big( \mathrm{CLAS}], \cdots, [\mathrm{CLAS}], \cdots, [\mathrm{CLAS}]\big)$} (see Fig.~\ref{fig:trajectory_recovery}). The number of pairs matches the number of inserted {\small $[\mathrm{MASK}]$} tokens, with each pair corresponding to a specific {\small $[\mathrm{MASK}]$}. The {\em General Task Head} decodes these task placeholders into the ID of segments for the generated ST-units at the positions of {\small $[\mathrm{MASK}]$} (see Sec.~\ref{sec:decoder}).
\end{itemize}
The task-oriented prompts structured by this template serve as the final inputs to the backbone model in \name.

\subsection{LLM-based Backbone Model}~\label{sec:backbone}
We employ a tunable LLM as the backbone model for our framework. In the implementation, the backbone LLM is \backbone~\cite{radford2019language}, but it can be replaced with other models based on user requirements. To enable efficient fine-tuning, we incorporate Low-Rank Adaptation (LoRA) modules into the backbone model. LoRA is a lightweight method for fine-tuning LLMs~\cite{hu2021loralowrankadaptationlarge}. It extends the backbone model parameters by attaching low-rank matrices externally. During fine-tuning, LoRA keeps the original model weights frozen and updates the model using low-rank matrix decomposition. This approach significantly reduces storage and computational costs while maintaining flexibility for task-specific adaptations.

In our model, LoRA modules are integrated into the {\em query}, {\em key}, and {\em value} matrices, as well as the FFN layers of each transformer block in \backbone. During training, only the parameters of the LoRA modules are updated, while the original parameters of \backbone remain frozen. This approach allows us to transfer \backbone's general sequence modeling capabilities to ST-unit sequence analysis tasks efficiently. For each input prompt {\small $\bm{\mathcal{X}}$}, the backbone processes the input as follows:
\begin{equation}\label{eq:LLM}\small
   \big\{\bm{Z}, \bm{V} \big\} = \mathrm{LLM}\Big(\bm{\mathcal{X}}, \; \bm{\Phi}_{\mathrm{LoRA}}\Big),
\end{equation}
where {\small $\bm{\Phi}_{\mathrm{LoRA}}$} is the tuneable parameters of the LoRA modules.

In Eq.~\eqref{eq:LLM}, the output sequence is divided into two parts: $i$) {\small $\bm{Z}$}, which is named as {\em output tokens} corresponding to the inputs task tokens {\small $\bm{X}^{\mathrm{(tsk)}}$}; and $ii$) {\small $\bm{V}$}, which corresponds to the remaining parts of the inputs prompt tokens. The output module (\ie {\em General Task Heads}) processes only the output tokens {\small $\bm{Z}$}, where the $k$-th token {\small $\bm{z}_{k}$} in {\small $\bm{Z}$} directly corresponds to the $k$-th {\small $[\mathrm{CLAS}]$} or {\small $[\mathrm{REG}]$} in the task placeholders.

%, \ie {\small $\big( \bm{X}^{\mathrm{(txt)}}, \bm{X}^{\mathrm{(st)}} \big)$}

\subsection{General-task Heads}~\label{sec:decoder}
\vspace{-0.3cm}

Unlike many ST representation learning methods that employ different task-specific output heads for various tasks, our model utilizes unified general-task heads to decode the output tokens {\small $\bm{Z}$} into results for different types of tasks. Similar to task tokens, the result tokens {\small $\bm{z}_k \in \bm{Z}$} are classified into two types: {\small $\bm{z}_k^{(\mathrm{clas})}$} corresponding to {\small $[\mathrm{CLAS}]$}, and {\small $\bm{z}_k^{(\mathrm{reg})}$} corresponding to {\small $[\mathrm{REG}]$}. We use multilayer perceptrons (MLPs) as decoders to map the result tokens to classification or regression outputs, as follows:
\begin{equation}\label{eq:mlp_decoder}\small
\begin{aligned}
    \;\;\;\;& \hat{\bm{y}}_k^{(\mathrm{clas})} = \mathrm{MLP}_c\left(\bm{z}_k^{(\mathrm{clas})}\right), \\
    \hat{y}_k^{(\mathrm{tim})} = \mathrm{MLP}_t&\left(\bm{z}_k^{(\mathrm{reg})}\right), \;\;%\\
    \hat{\bm{y}}_k^{(\mathrm{reg})} = \mathrm{MLP}_r\left(\bm{z}_k^{(\mathrm{reg})}\right),
\end{aligned}
\end{equation}
Here, {\small $\mathrm{MLP}_c(\cdot)$} generates results for classification tasks, {\small $\mathrm{MLP}_t(\cdot)$} handles timestamp regression (used in tasks such as TTE and timestamp generation for trajectory recovery), and {\small $\mathrm{MLP}_r(\cdot)$} is responsible for regressing other types of outputs. The predicted classification, timestamp, and general regression results are denoted by {\small $\hat{\bm{y}}_k^{(\mathrm{clas})}$} (in one-hot encoding), {\small $\hat{y}_k^{(\mathrm{tim})}$}, and {\small $\hat{\bm{y}}_k^{(\mathrm{reg})}$}, respectively. Given the significant differences between temporal and spatial features, a specialized MLP decoder, {\small $\mathrm{MLP}_t(\cdot)$}, is used exclusively for timestamp regression. % to enhance performance.
%the temporal feature have

\paratitle{\em Remark:} Guided by the textual instructions in task-oriented prompts, our model is equipped with the knowledge of which task to execute. By adapting to different instructions, the backbone model and general-task heads generate task-specific outputs for various types of tasks, effectively addressing the challenge of adapting to diverse spatiotemporal analysis tasks (Challenge 2). Furthermore, we leverage a LLM as the backbone, offering two key advantages: First, the LLM's powerful text processing capabilities enable our model to accurately interpret instructions and adapt the data processing accordingly. Second, the LLM's strong general sequence modeling abilities can be seamlessly transferred to ST sequence data analysis tasks, enhancing the model's performance and versatility.

%% file: content/wang_version/3_4_training.tex
\begin{figure}[t]
     \centering
     \includegraphics[width=1.0\linewidth]{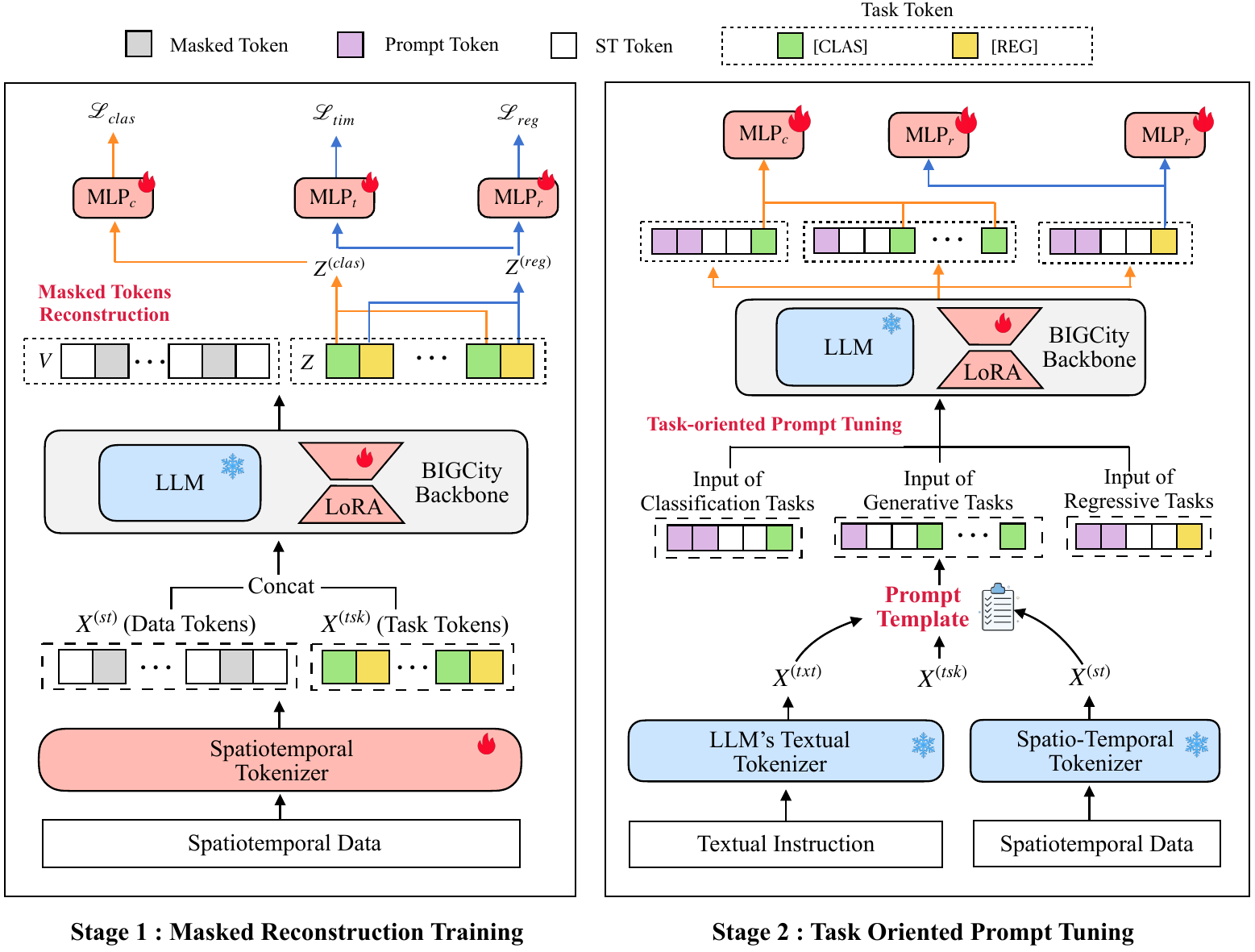}
     \caption{The hierarchical training strategy of \name.}
     \label{fig:training_framework}
     % \vspace{-0.3cm}
\end{figure}

\section{Model Training}
\label{sec:training}

As shown in Fig.~\ref{fig:training_framework}, \name adopts a two-stage training strategy comprising {\em Masked Reconstruction Training} and {\em Multi-Task Prompt Tuning}. In the first stage, masked reconstruction training, a self-supervised approach is employed to enable the model to learn general spatiotemporal dependencies from ST sequence data. In the second stage, multi-task prompt tuning, a multi-task co-training method is utilized to allow the model to handle a variety of tasks within a unified framework, eliminating the need for task-specific fine-tuning.

\subsection{Masked Reconstruction Training}

In this stage, we employ a masked reconstruction task to enable the model to capture general features of ST sequences without being tailored to specific data analysis tasks. Masked reconstruction is a widely used approach in LLM pre-training, where certain samples from the input sequence are randomly masked, and the model predicts the masked samples. This task allows the model to learn intrinsic correlations within the input sequence data without requiring labeled data. The specific process of masked reconstruction training in our model is described as follows.

\paratitle{Inputs.} Given an ST-unit sequence corresponding to a trajectory or traffic state series, \ie {\small $\mathcal{U} = (\bm{U}_{1}, \cdots, \bm{U}_{L})$}, the ST tokenizer encodes it into an ST token sequence {\small $\bm{X}^{(\mathrm{st})} = (\bm{x}_{1}, \ldots, \bm{x}_{L})$}. We randomly mask $K$ tokens in {\small $\bm{X}^{(\mathrm{st})}$}, replacing them with the mask token {\small $[\mathrm{MASK}]$}, resulting in {\small $\tilde{\bm{X}}^{(\mathrm{st})} = (\bm{x}_{1}, \cdots, [\mathrm{MASK}]_1, \cdots, \bm{x}_{l}, \cdots, [\mathrm{MASK}]_K, \cdots, \bm{x}_{L})$}. For each {\small $[\mathrm{MASK}]_k$}, we assign a task placeholder pair {\small $([\mathrm{CLAS}]_k, [\mathrm{REG}]_k)$}, producing a task token sequence:
\begin{equation}\label{eq:task_pretrain}\small
  \bm{X}^{(\mathrm{tsk})} = \left(\big(\bm{x}_1^{(\mathrm{clas})}, \bm{x}_1^{(\mathrm{reg})}\big), \ldots, \big(\bm{x}_K^{(\mathrm{clas})}, \bm{x}_K^{(\mathrm{reg})}\big)\right).
\end{equation}
Finally, the token sequence {\small $\bm{\mathcal{X}} = \big( \tilde{\bm{X}}^{(\mathrm{st})}, \bm{X}^{(\mathrm{tsk})} \big)$} is fed into the backbone LLM model for masked reconstruction training.

\paratitle{Outputs.} Given {\small $\bm{\mathcal{X}}$} as input to the backbone model, the output is a result token sequence {\small $\bm{Z}$}, where each element corresponds one-to-one with elements in {\small $\bm{X}^{(\mathrm{tsk})}$}, as follows:
\begin{equation}\label{eq:result_tokens}\small
  \bm{Z} = \left(\big(\bm{z}_1^{(\mathrm{clas})}, \bm{z}_1^{(\mathrm{reg})}\big), \ldots, \big(\bm{z}_K^{(\mathrm{clas})}, \bm{z}_K^{(\mathrm{reg})}\big)\right).
\end{equation}
For the $k$-th output token {\small $\big(\bm{z}_k^{(\mathrm{clas})}, \bm{z}_k^{(\mathrm{reg})}\big)$}, the general-task head (Eq.~\eqref{eq:mlp_decoder}) decodes it into an predicted ST-unit:
\begin{equation}\label{eq:predicted_ST-unit}\small
  \hat{\mathcal{U}}_k = \left(\hat{\bm{y}}_k^{(\mathrm{clas})}, \hat{\bm{y}}_k^{(\mathrm{reg})},\hat{y}_k^{(\mathrm{tim})}\right),
\end{equation}
which represents the reconstructed sample for the ST-unit masked by {\small $[\mathrm{MASK}]_k$}. In Eq.~\eqref{eq:predicted_ST-unit}, {\small $\hat{\bm{y}}_k^{(\mathrm{clas})}$} is the predicted road segment ID, {\small $\hat{\bm{y}}_k^{(\mathrm{reg})}$} is the predicted dynamic feature of the road segment, and {\small $\hat{y}_k^{(\mathrm{tim})}$} is the predicted timestamp.

\paratitle{Loss Function.} We employ the Cross-Entropy loss for classification and the Mean Squared Error (MSE) loss for regression. For each masked ST-unit, there are three losses:
\begin{equation}\footnotesize
\begin{aligned}
    &\;\;\;\mathcal{L}_{\mathrm{clas}}  = \sum_{k=1}^{K} \bm{\phi}\Big(r_{k}\Big) \log\left(\hat{\bm{y}}_k^{(\mathrm{clas})}\right), \\
    \mathcal{L}_{\mathrm{reg}}   = \sum_{k=1}^{K}& \left\| \bm{e}^{(d)}_{k} - \hat{\bm{y}}_k^{(\mathrm{reg})}\right\|^2_F,\;\;
    \mathcal{L}_{\mathrm{tim}}   = \sum_{k=1}^{K} \left(\tau_k - \hat{y}_k^{(\mathrm{tim})}\right)^2,
\end{aligned}
\end{equation}
where {\small $\bm{\phi}\big(r_{k}\big)$}, {\small $\bm{e}^{(d)}_{k}$}, and {\small $\tau_k$} are the ground truths for {\small $\hat{\mathcal{U}}_k$}. The final loss function for masked reconstruction training is:
\begin{equation}\label{pre_loss}\small
    \mathcal{L}_{\mathrm{MRT}} = \frac{1}{K \times N} \sum_{n=1}^{N} \left( \mathcal{L}_{\mathrm{clas}}^{(n)} +
     \lambda_1 \mathcal{L}_{\mathrm{reg}}^{(n)} + \lambda_2 \mathcal{L}_{\mathrm{tim}}^{(n)} \right),
\end{equation}
where {\small $\lambda_1$} and {\small $\lambda_2$} are predefined parameters, the superscript {\small $(n)$} indicates the index of a training sample, and {\small $N$} is the total number of training samples.

During Masked Reconstruction Training, the ST tokenizer and the LoRA module in the backbone are jointly trained by minimizing the loss function {\small $\mathcal{L}_{\mathrm{MRT}}$}.

\subsection{Task-oriented Prompt Tuning}~\label{sec:prompt_tuning}
This stage trains our model to perform various ST data analysis tasks using task-oriented prompts.

\paratitle{Inputs.} In this stage, the inputs are task-oriented prompts for diverse tasks. Datasets for all task types are combined into a unified training dataset, referred to as the {\em full training set}. For each sample in the full training set, a corresponding prompt is generated as {\small $\bm{\mathcal{X}} = \big(\bm{X}^{(\mathrm{txt})}, \bm{X}^{(\mathrm{st})}, \bm{X}^{(\mathrm{tsk})}\big)$}, which serves as the input to the backbone model of \name.

In the implementation, we incorporate three types of tasks for prompt tuning: {\em classification}, {\em regression}, and {\em generation}. The textual instruction tokens {\small $\bm{X}^{(\mathrm{txt})}$} in {\small $\bm{\mathcal{X}}$} follow fixed templates as shown in Fig.~\ref{fig:prompt_example}. The data tokens {\small $\bm{X}^{(\mathrm{st})}$} and task tokens {\small $\bm{X}^{(\mathrm{tsk})}$} for these tasks are described as follows:
\begin{itemize}[leftmargin=*]
\item {\em Classification Tasks} include next hop prediction and trajectory classification. Trajectory classification tasks further comprise trajectory traffic pattern classification and user-trajectory linkage. The data tokens {\small $\bm{X}^{(\mathrm{st})}$} in {\small $\bm{\mathcal{X}}$} represent a sequence of ST-units corresponding to a trajectory. The task tokens {\small $\bm{X}^{(\mathrm{tsk})}$} consist of a single {\small $[\mathrm{CLAS}]$}, corresponding to the class label to be predicted.
\item {\em Regression Tasks} include travel time estimation and traffic state prediction. $i$) For the travel time estimation task, the data tokens {\small $\bm{X}^{(\mathrm{st})}$} are a trajectory ST-unit sequence, where timestamps at locations with unknown travel times are replaced by {\small $[\mathrm{MASK}]$}. The task tokens are a sequence of {\small $[\mathrm{REG}]$}, each corresponding to an unknown timestamp. $ii$) For traffic state prediction, the data tokens represent the first half of a traffic state sequence, and the task tokens are a sequence of {\small $[\mathrm{REG}]$}, each corresponding to the rest traffic state steps to be predicted.
\item {\em Generation Task} includes trajectory recovery. The data tokens consist of a low-sampling-rate trajectory where locations to be recovered are inserted with {\small $[\mathrm{MASK}]$}. The task tokens are a sequence of {\small $[\mathrm{CLAS}]$} with the same number of elements as the {\small $[\mathrm{MASK}]$} tokens. Each {\small $[\mathrm{CLAS}]$} corresponds to a sample to be recovered in the low-sampling-rate trajectory.
\end{itemize}

\paratitle{Outputs.} For classification tasks, the general-task heads decode the output token corresponding to {\small $[\mathrm{CLAS}]$} into the predicted class label. For regression tasks, the output token corresponding to {\small $[\mathrm{REG}]$} is decoded into predicted timestamps or traffic states. For generation tasks, the output tokens corresponding to {\small $[\mathrm{CLAS}]$} are decoded into the road segment IDs of the recovered samples. %, while those corresponding to {\small $[\mathrm{REG}]$} are decoded into the timestamps and traffic states of the recovered samples.

\paratitle{Loss Function.} In task-oriented prompt tuning, we use samples from all tasks, \ie the full training set, to co-train the model. The overall loss function is defined as:
\begin{equation}\label{}
    \mathcal{L}_{\mathrm{PT}} = \mathcal{L}_{\mathrm{CLAS}} + \lambda_2 \mathcal{L}_{\mathrm{REG}} + \lambda_3 \mathcal{L}_{\mathrm{GEN}},
\end{equation}
where {\small $\mathcal{L}_{\mathrm{CLAS}}$}, {\small $\mathcal{L}_{\mathrm{REG}}$}, and {\small $\mathcal{L}_{\mathrm{GEN}}$} represent the loss functions for classification, regression, and generation tasks, respectively. Cross-entropy loss is applied for classification tasks, while mean squared error (MSE) loss is used for regression tasks.

In task-oriented prompt tuning, only the parameters in the LoRA modules of the backbone model are updated using the loss function {\small $\mathcal{L}_{\mathrm{PT}}$}, while the parameters of the ST tokenizer remain frozen. After completing both the masked reconstruction training and task-oriented prompt tuning, the ST tokenizer and the \submodel are assembled to form the complete \name model, capable of ST data analysis across multiple tasks in multiple data modalities ({\em MTMD}).

%% file: content/4_experiments.tex
\begin{table}[htbp]
\captionsetup{font=small}
  \centering
  \renewcommand{\arraystretch}{1.0}
  \caption{Types of ST Data Analysis Tasks}
    \begin{tabular}{c|c|c}
    \toprule
    \bf{Task Type} & \bf{Trajectory Based} & \bf{Traffic State Based} \\
    \midrule
%    \midrule
    \multirow{2}[4]{*}{{\it Classification}} & Next Hop Prediction & \multirow{2}[4]{*}{\textbackslash} \\
\cmidrule{2-2}          & \shortstack{Trajectory Classification } &   \\
    \midrule
    \multirow{2}[4]{*}{{\it Regression}} & \multirow{2}[4]{*}{Travel Time Estimation } & One-Step Prediction \\
\cmidrule{3-3}          &     & Multi-Step Prediction \\
    \midrule
    {\it Comparison} & Most Similar Search  &  \textbackslash \\
    \midrule
    {\it Generation} & Trajectory Recovery  & Traffic State Imputation \\
    \bottomrule
    \end{tabular}%
   % \vspace{-0.1cm}
  \label{tab:task_types}
\end{table}%

\begin{table}[t]
\captionsetup{font=small}
\fontsize{8.5pt}{9pt}\selectfont
  \centering
  \renewcommand{\arraystretch}{1.25}
  \caption{Statistics of the Three Datasets.}
  \vspace{-0.1cm}
    \begin{tabular}{c|c|c|c}
    \toprule[0.9pt]
    \bf{Dataset}        & \bj        & \xa        & \cd        \\ \hline
    \hline
    Time Span      & one month & one month & one month \\
    Trajectory     & 1018312   & 384618    & 559729    \\
    User Number    & 1677      & 26787     & 48295     \\
    Road Segments  & 40306     & 5269      & 6195      \\
    \bottomrule[0.9pt]
    \end{tabular}%
    % \vspace{-0.5cm}
  \label{tab:data_detail}%
\end{table}

\begin{table*}[t]
\captionsetup{font=small}
  \renewcommand{\arraystretch}{1}
  \centering
  \caption{\centering Performance on the trajectory-based non-generative tasks, \ie Travel Time Estimation, Trajectory Classification, Trajectory Next Hop Prediction, Most Similar Trajectory Search. Particularly, MAPE here are percentage numbers.}
    \resizebox{\textwidth}{!}{
    \begin{threeparttable}
    \begin{tabular}{c|c|ccc|ccc|ccc|ccc}
    \toprule
    \multicolumn{2}{c|}{\textbf{Task}} & \multicolumn{3}{c|}{\textbf{Travel Time Estimation}} & \multicolumn{3}{c|}{\textbf{Trajectory Classification}} & \multicolumn{3}{c|}{\textbf{Next Hop Prediction}} & \multicolumn{3}{c}{\textbf{Most Similar Search}} \\
    \midrule
    Data & \multicolumn{1}{c|}{Model} & MAE$\downarrow$ & RMSE$\downarrow$ & MAPE$\downarrow$ & ACC$\uparrow$ & F1$\uparrow$ & AUC$\uparrow$ & ACC$\uparrow$ & MRR@5$\uparrow$ & NDC@5 $\uparrow$ & HR@1$\uparrow$ & HR@5$\uparrow$ & HR@10$\uparrow$ \\
    \midrule
    \midrule
    \multirow{8}[2]{*}{\textbf{BJ}} & Tr2v  & 10.13 & 56.83 & 37.95 & 0.811 & 0.852 & 0.837 & 0.633 & 0.746 & 0.784 & 0.607 & 0.766 & 0.867 \\
          & T2v   & 10.03 & 56.65 & 36.42 & 0.814 & 0.863 & 0.879 & 0.623 & 0.731 & 0.769 & \textcolor{black!70}{\uline{0.788}} & \textcolor{black!70}{\uline{0.885}} & \textcolor{black!70}{\uline{0.932}} \\
          & TBR   & 9.981  & 36.97 & 34.25  & 0.818 & 0.876  & 0.871 & 0.537  & 0.625 & 0.659 & 0.396 & 0.499 & 0.538 \\
          & Toa   & 10.79 & 57.41 & 35.37 & 0.821  & 0.861 & 0.862 & 0.714 & 0.841 & 0.878 & 0.326 & 0.399 & 0.767 \\
          & JCL   & 10.23 & 46.49 & 41.22 & 0.808 & 0.859 & 0.873 & 0.714 & 0.841 & 0.881  & 0.531 & 0.699 & 0.936 \\
          & STA   & \textcolor{black!70}{\uline{9.156}}  & \textcolor{black!70}{\uline{35.41}} & \textcolor{black!70}{\uline{32.01}} & \textcolor{black!70}{\uline{0.853}} & \textcolor{black!70}{\uline{0.885}} & \textcolor{black!70}{\uline{0.905}} & 0.734 & 0.836 & 0.882 & 0.776 & 0.878 & 0.934 \\
          & JRM   & 10.18 & 41.88 & 39.51 & 0.849 & 0.879 & 0.897 & \textcolor{black!70}{\uline{0.746}} & \textcolor{black!70}{\uline{0.843}} & \textcolor{black!70}{\uline{0.896}} & 0.681 & 0.852 & 0.906 \\
          \rowcolor{gray!10} & {\bf Ours} & \textcolor{black}{\bf 8.869}  & \textcolor{black}{ \bf 33.21} & \textcolor{black}{\bf 30.34} & \textcolor{black}{\bf 0.872} & \textcolor{black}{ \bf 0.891} & \textcolor{black}{\bf 0.909} & \textcolor{black}{\bf 0.751} & \textcolor{black}{ \bf 0.855} & \textcolor{black}{\bf 0.902} & \textcolor{black}{\bf 0.801} & \textcolor{black}{ \bf 0.895} & \textcolor{black}{\bf 0.952} \\
    \midrule
    \midrule
    Data & \multicolumn{1}{c|}{Model} & MAE$\downarrow$ & RMSE$\downarrow$ & MAPE$\downarrow$ & Mi-F1$\uparrow$ & Ma-F1$\uparrow$ & Ma-Re$\uparrow$ & ACC$\uparrow$ & MRR@5$\uparrow$ & NDC@5$\uparrow$ & HR@1$\uparrow$ & HR@5$\uparrow$ & HR@10$\uparrow$ \\
    \midrule
    \multirow{8}[2]{*}{\textbf{XA}} & Tr2v  & 2.051  & 3.147  & 35.14 & 0.086 & 0.085 & 0.093 & 0.679 & 0.759 & 0.788 & 0.673 & 0.854 & 0.889 \\
          & T2v   & 2.035  & 3.132  & 33.73 & 0.086 & 0.082 & 0.089 & 0.672 & 0.747 & 0.774 & 0.733 & 0.821 & 0.877 \\
          & TBR   & 2.016  & 3.121  & 32.13 & 0.091 & 0.088 & 0.081 & 0.568 & 0.633 & 0.657 & 0.538 & 0.67  & 0.725 \\
          & Toa   & 2.152  & 3.266  & 33.93 & 0.099 & 0.095 & 0.092 & 0.778 & 0.887 & 0.913 & 0.283  & 0.393 & 0.442 \\
          & JCL   & 2.173  & 3.257  & 33.12 & 0.093 & 0.091 & 0.095 & 0.793 & 0.889 & 0.919 & 0.335 & 0.551  & 0.634 \\
          & STA   & \textcolor{black!70}{\uline{1.833}}  & \textcolor{black!70}{\uline{2.982}}  & \textcolor{black!70}{\uline{30.57}} & \textcolor{black!70}{\uline{0.101}} & \textcolor{black!70}{\uline{0.098}} & \textcolor{black!70}{\uline{0.102}} & 0.825 & 0.903 & 0.928 & \textcolor{black!70}{\uline{0.741}} & \textcolor{black!70}{\uline{0.883}}  & \textcolor{black!70}{\uline{0.893}} \\
          & JRM   & 1.915  & 3.152  & 31.88 & 0.097 & 0.094 & 0.097 & \textcolor{black}{\uline{0.829}} & \textcolor{black!70}{\uline{0.906}} & \textcolor{black!70}{\uline{0.934}} & 0.703 & 0.826 & 0.863 \\
          \rowcolor{gray!10} & {\bf Ours} & \textcolor{black}{\bf 1.723}  & \textcolor{black}{\bf 2.614}  & \textcolor{black}{\bf 29.76} & \textcolor{black}{\bf 0.112}  & \textcolor{black}{\bf 0.104} & \textcolor{black}{\bf 0.113} & \textcolor{black}{\bf 0.837} & \textcolor{black}{\bf 0.923} & \textcolor{black}{\bf 0.942}  & \textcolor{black}{\bf 0.791} & \textcolor{black}{\bf 0.887} & \textcolor{black}{\bf 0.909} \\
    \midrule
    \midrule
    Data & \multicolumn{1}{c|}{Model} & MAE$\downarrow$ & RMSE$\downarrow$ & MAPE$\downarrow$ & Mi-F1$\uparrow$ & Ma-F1$\uparrow$ & Ma-Re$\uparrow$ & ACC$\uparrow$ & MRR@5$\uparrow$ & NDC@5$\uparrow$ & HR@1$\uparrow$ & HR@5$\uparrow$ & HR@10$\uparrow$ \\
    \midrule
    \multirow{8}[2]{*}{\textbf{CD}} & Tr2v  & 1.635  & 2.432  & 34.74 & 0.142 & 0.152  & 0.156 & 0.726 & 0.809 & 0.837 & 0.607  & 0.748 & 0.794 \\
          & T2v   & 1.632  & 2.433  & 34.45 & 0.149 & 0.151 & 0.158 & 0.711 & 0.791 & 0.819 & 0.543 & 0.715   & 0.753 \\
          & TBR   & 1.620  & 2.405  & 34.15 & 0.142 & 0.156 & 0.155 & 0.608 & 0.682 & 0.708 & 0.409 & 0.532  & 0.576 \\
          & Toa   & 1.708  & 2.493  & 37.23  & 0.143  & 0.152 & 0.153 & 0.789 & 0.872 & 0.911 & 0.225 & 0.322 & 0.357 \\
          & JCL   & 1.657  & 2.481  & 36.42 & 0.148 & 0.164 & 0.157 & 0.792 & 0.881 & 0.904 & 0.348 & 0.552 & 0.642 \\
          & STA   & 1.433  & 2.394  & 32.12 & \textcolor{black!70}{\uline{0.151}} & \textcolor{black!70}{\uline{0.163}} & \textcolor{black!70}{\uline{0.159}} & 0.795 & 0.885 & 0.919 & 0.607 & 0.757 & 0.776 \\
          & JRM   & \textcolor{black!70}{\uline{1.372}}  & \textcolor{black!70}{\uline{2.253}}  & \textcolor{black!70}{\uline{30.79}} & 0.143 & 0.152 & 0.151 & \textcolor{black!70}{\uline{0.798}} & \textcolor{black!70}{\uline{0.888}} & \textcolor{black!70}{\uline{0.927}} & \textcolor{black!70}{\uline{0.631}} & \textcolor{black!70}{\uline{0.774}} & \textcolor{black!70}{\uline{0.815}} \\
          \rowcolor{gray!10} & {\bf Ours} & \textcolor{black}{\bf 1.287}  & \textcolor{black}{\bf 2.181}  & \textcolor{black}{\bf 28.59} & \textcolor{black}{\bf 0.153} & \textcolor{black}{\bf 0.169} & \textcolor{black}{\bf 0.162} & \textcolor{black}{\bf 0.821} & \textcolor{black}{\bf 0.912}  & \textcolor{black}{\bf 0.938} & \textcolor{black}{\bf 0.646} & \textcolor{black}{\bf 0.787} & \textcolor{black}{\bf 0.821} \\
    \bottomrule
    \end{tabular}%
    \vspace{0.2cm}
    \begin{tablenotes}
         \footnotesize
         \item[*] The bold results are the best, and the underlined results are the second best. The metric with "$\uparrow$" ("$\downarrow$") means that a larger (smaller) result is better.
    \end{tablenotes}
    \end{threeparttable}
    }
   \label{tab:TBNG_task}%
\end{table*}%

\begin{table*}[t]
\captionsetup{font=small}
\fontsize{13pt}{15pt}\selectfont
\renewcommand{\arraystretch}{1.25}
  \centering
  \caption{Performance on trajectory based generative tasks, \ie Trajectory Recovery. $85\%$, $90\%$, and $95\%$ are masked ratios.}
  \resizebox{\textwidth}{!}{
    \begin{tabular}{c|ccc|ccc|ccc|ccc|ccc|ccc}
    \toprule[1.2pt]
    {\bf Metric} & \multicolumn{9}{c|}{{\bf Accuracy $\uparrow$}} & \multicolumn{9}{c}{{\bf Macro-F1 $\uparrow$}} \\
    \midrule
    \multirow{2}{*}{\diagbox{{\bf Models}}{{\bf Data}}} & \multicolumn{3}{c|}{\textbf{BJ}} & \multicolumn{3}{c|}{\textbf{XA}} & \multicolumn{3}{c|}{\textbf{CD}} & \multicolumn{3}{c|}{\textbf{BJ}} & \multicolumn{3}{c|}{\textbf{XA}} & \multicolumn{3}{c}{\textbf{CD}} \\
     & 85\%  & 90\%  & 95\%  & 85\%  & 90\%  & 95\%  & 85\%  & 90\%  & 95\%  & 85\%  & 90\%  & 95\%  & 85\%  & 90\%  & 95\%  & 85\%  & 90\%  & 95\% \\
    \midrule
    \midrule
    Linear+HMM & 0.219 & 0.205 & 0.196 & 0.275 & 0.239 & 0.207 & 0.289 & 0.268 & 0.233
               & 0.098 & 0.095 & 0.089 & 0.125 & 0.101 & 0.094 & 0.131 & 0.117 & 0.099  \\
    DTHR+HMM & 0.236 & 0.227 & 0.208 & 0.269 & 0.218 & 0.201 & 0.296 & 0.264 & 0.224
             & 0.118 & 0.097 & 0.091 & 0.135 & 0.121 & 0.105 & 0.141 & 0.119 & 0.106  \\
    MTrajRec & 0.456 & 0.418 & 0.323 & 0.495 & 0.443 & 0.338 & 0.512 & 0.459 & 0.347
             & 0.201 & 0.177 & 0.136 & 0.221 & 0.199 & 0.145 & 0.244 & 0.240 & 0.164   \\
    RNTrajRec & {\textcolor{black!70}{\ul 0.475}} & {\textcolor{black!70}{\ul 0.439}} & {\textcolor{black!70}{\ul 0.338}} & {\textcolor{black!70}{\ul 0.503}} & {\textcolor{black!70}{\ul 0.456}} & { \textcolor{black!70}{\ul 0.359}} & {\textcolor{black!70}{\ul 0.523}}
    & {\textcolor{black!70}{\ul 0.478}} & {\textcolor{black!70}{\ul 0.369}}
    & {\textcolor{black!70}{\ul 0.205}} & {\textcolor{black!70}{\ul 0.181}} & {\textcolor{black!70}{\ul 0.152}}   & {\textcolor{black!70}{\ul 0.267}}   & {\textcolor{black!70}{\ul 0.226}}   & { \textcolor{black!70}{\ul 0.173}}    & {\textcolor{black!70}{\ul 0.292}}   & {\textcolor{black!70}{\ul 0.257}}   & {\textcolor{black!70}{\ul 0.185}}   \\\midrule
    \rowcolor{gray!10} \textbf{Ours} & \textbf{\textcolor{black}{0.518}} & \textbf{\textcolor{black}{0.471}} & \textbf{\textcolor{black}{0.368}} & \textbf{\textcolor{black}{0.562}} & \textbf{\textcolor{black}{0.489}} & \textbf{\textcolor{black}{0.381}} & \textbf{\textcolor{black}{0.585}}   & \textbf{\textcolor{black}{0.513}} & \textbf{\textcolor{black}{0.405}}        & \textbf{\textcolor{black}{0.259}}      & \textbf{\textcolor{black}{0.217}}      & \textbf{\textcolor{black}{0.177}}      & \textbf{\textcolor{black}{0.309}}   & \textbf{\textcolor{black}{0.258}}       & \textbf{\textcolor{black}{0.194}}
     & \textbf{\textcolor{black}{0.321}}       & \textbf{\textcolor{black}{0.269}}      & \textbf{\textcolor{black}{0.212}} \\
    \bottomrule[1.2pt]
    \end{tabular}%
   }
   % \vspace{-0.2cm}
  \label{tab:traj_recovery}%
\end{table*}%

\begin{table}[t]\footnotesize
\captionsetup{font=small}
\tabcolsep=0.07cm
% \setlength{\tabcolsep}{0.7pt}
%\fontsize{5.6pt}{6pt}\selectfont
\renewcommand{\arraystretch}{1.2}
  \centering
  \caption{The Performance in Traffic State Tasks}
  %\resizebox{\textwidth}{!}
  {
    \begin{tabular}{c|ccc|ccc|ccc}
    \toprule[0.6pt]
       {\bf{Data}} & \multicolumn{9}{c}{\xa} \\ \midrule[0.2pt]
     Task  & \multicolumn{3}{c|}{\bf{One-Step}} & \multicolumn{3}{c|}{\bf{Multi-Step}} & \multicolumn{3}{c}{\bf{Imputation}} \\    \midrule[0.2pt]
           {Metric} & MAE & MAPE & RMSE & MAE & MAPE & RMSE & MAE & MAPE & RMSE\\
    \midrule[0.2pt]
     DCR & 1.092  & 11.77  & 2.312 & 1.293  & 16.38  & 2.492  & \textcolor{black!70}{{\ul 0.585}}  & \textcolor{black!70}{{\ul 7.493}}  & \textcolor{black!70}{{\ul 1.403}} \\
           GWN & 1.113  & 11.44  & 2.264 & 1.304  & 15.59  & 2.331 & 0.847 & 10.63 & 1.833 \\
           MTG & 1.072  & 10.56  & 1.903  & 1.223  & 14.91 & \textcolor{black!70}{{\ul 2.163 }} & 0.906 & 11.12 & 1.790 \\
           TrG & 1.103  & 11.46  & 2.042 & 1.263 & 15.90 & 2.423 & 0.944 & 11.79 & 1.815 \\
           STG & 1.122  & 12.59  & 2.272 & 1.392 & 17.34 & 2.304 & 0.989 & 12.40 & 1.709 \\
           STN & 0.974  & 10.27  & 1.973 & 1.268 & 15.64 & 2.281 & 0.940 & 11.64 & 1.789 \\
           SST & \textcolor{black!70}{{\ul 0.802}}  & \textcolor{black!70}{{\ul 9.972}}  & \textcolor{black!70}{{\ul 1.873}} & \textcolor{black!70}{{\ul 1.183 }} & \textcolor{black!70}{{\ul 14.21}} & 2.292 & 0.883 & 11.23 & 1.736 \\
          \midrule[0.2pt]
          \rowcolor{gray!10} \bf{Ours}  &  \textcolor{black}{{\bf 0.791}}   & \textcolor{black}{{\bf 9.732}}    &\textcolor{black}{{\bf 1.743}} &  \textcolor{black}{{\bf 1.162}}   & \textcolor{black}{{\bf 14.01}}  & \textcolor{black}{{\bf 2.143}}
          &  \textcolor{black}{{\bf 0.536}}   & \textcolor{black}{{\bf 6.671}}  & \textcolor{black}{{\bf 1.335}} \\
          \midrule[0.2pt]
             {\bf{Data}} & \multicolumn{9}{c}{\xa} \\\midrule[0.2pt]
     Task  & \multicolumn{3}{c|}{\bf{One-Step}} & \multicolumn{3}{c|}{\bf{Multi-Step}} & \multicolumn{3}{c}{\bf{Imputation}} \\    \midrule[0.2pt]
           {Metric} & MAE & MAPE & RMSE & MAE & MAPE & RMSE & MAE & MAPE & RMSE\\
    \midrule[0.2pt]
        %\multicolumn{10}{c}{\cd} \\
        DCR & 1.232  & 13.02  & 2.324 & 1.552 & 18.22 & 2.862 & \textcolor{black!70}{{\ul 0.731}}  & \textcolor{black!70}{{\ul 8.601}}  & \textcolor{black!70}{{\ul 1.704}} \\
           GWN & 1.342  & 12.71  & 2.414  & 1.612 & 18.14 & 2.713 & 1.024 & 11.80 & 2.121 \\
           MTG & 1.183  & 11.93  & \textcolor{black!70}{{\ul 2.132}} & 1.413 & \textcolor{black!70}{{\ul 16.76}} & \textcolor{black!70}{{\ul 2.506}} & 1.109 & 12.71 & 2.055 \\
           TrG & 1.234  & 12.39  & 2.312 & 1.562 & 17.69 & 2.761 & 1.165 & 13.14 & 2.082 \\
           STG & 1.352  & 12.91  & 2.413 & 1.633 & 18.77 & 2.603 & 1.235 & 14.37 & 2.046 \\
           STN & 1.203  & 11.99  & 2.201 & 1.491 & 17.02 & 2.594 & 1.152 & 12.88 & 2.061 \\
           SST & \textcolor{black!70}{{\ul 1.163}}  & \textcolor{black!70}{{\ul 11.77}}  & 2.191  & \textcolor{black!70}{{\ul 1.452}} & 17.01 & 2.954 & 1.027 & 11.84 & 2.005 \\
          \midrule[0.2pt]
          \rowcolor{gray!10} \bf{Ours}  & \textcolor{black}{{\bf 1.122}}    & \textcolor{black}{{\bf 11.16}}    & \textcolor{black}{{\bf 2.103}} &\textcolor{black}{{\bf 1.412}}    & \textcolor{black}{{\bf 15.98}}    & \textcolor{black}{{\bf 2.471}}  &  \textcolor{black}{{\bf 0.665}}   & \textcolor{black}{{\bf 8.192}}  & \textcolor{black}{{\bf 1.617}} \\
    \bottomrule[0.6pt]
    \end{tabular}%
    }
    % \vspace{-0.2cm}
  \label{tab:traffic_prediction}%
\end{table}%

\vspace{0.05cm}
\section{Experiments}
\subsection{Experimental Setting}
\paratitle{Spatiotemporal (ST) Tasks.} As a multi-task model, \name is co-trained on four types of ST tasks, and simultaneously tackling eight specific tasks, as listed in Tab.~\ref{tab:task_types}. While current baselines demands individual training for each task.

\paratitle{Datasets.} We evaluated \name on three real-world datasets: Beijing (BJ), Xi'an (XA), and Chengdu (CD). The BJ dataset consists of taxi trajectories collected in November 2015~\cite{START}, while the XA and CD datasets include online car-hailing trajectories from November 2018, provided by the DiDi GAIA project\footnote{\url{https://www.didiglobal.com/news/newsDetail?id=199&type=news}}. Road networks for all three cities were extracted from OpenStreetMap (OSM)\footnote{\url{https://www.openstreetmap.org/}}, and trajectories were map-matched to the networks to compute traffic states. Each time slice for traffic states spans 30 minutes. Due to sparse trajectories in the BJ dataset, dynamic traffic state features from ST units (Eq.~\eqref{eq:traffic_series}) were excluded, a limitation common in trajectory datasets. For experiments, XA and CD datasets were split $6:2:2$ for training, validation, and testing, while BJ was split $8:1:1$. Dataset statistics are provided in Tab.~\ref{tab:data_detail}. In addition, our data are processed by libcity~\cite{wang2021libcity}

\paratitle{Baselines.}
\name is evaluated by comparing with seventeen task-specific baselines of each particular task. For classification, regression and comparison tasks of trajectory input, we select the seven current state-of-art (SOTA) trajectory representation models: {Trajectory2vec} (Tr2v)~\cite{trajectory2vec}, {T2vec} (T2v)~\cite{li2018deep}, {TremBR} (TBR)~\cite{Trembr}, {Toast} (Toa)~\cite{Toast}, {JCLRNT} (JCL)~\cite{JCLRNT}, {START} (STA)~\cite{jiang2023self}, and {JGRM} (JRM)~\cite{ma2024more}. For traffic state input, \name is compared with six current SOTA traffic state prediction models: DCRNN~\cite{li2017diffusion}, GWNET~\cite{wu2019graph}, MTGNN~\cite{wu2020connecting}, TrGNN~\cite{li2021traffic}, STGODE~\cite{fang2021spatial}, ST-Norm\cite{deng2021st}, and SSTBAN~\cite{guo2023self}. Table~\ref{tab:traffic_prediction}. For generation tasks, \name is compared with four recovery models: Linear+HMM~\cite{hoteit2014estimating}, DTHR+HMM, MTrajRec~\cite{MTrajRec}, and RNTrajRec~\cite{RNTrajRec}.

\paratitle{Evaluation Metrics.}
% \begin{itemize} [leftmargin=*]
% \item Trajectory Travel Time Estimation: we adopt three metrics, including mean absolute error (MAE), mean absolute percentage error (MAPE), and root mean square error (RMSE).
% \item Next Hop Prediction: Following the settings in~\cite{START}, we used three metrics: Accuracy (ACC), mean reciprocal rank at 5 (MRR@5) and Normalized Discounted Cumulative Gain at 5 (NDCG@5).
% \item Trajectory Classification: We use Accuracy (ACC), F1-score (F1), and Area Under ROC (AUC) to evaluate binary classification on BJ dataset. Using Micro-F1, Macro-F1 and Macro-Recall on XA and CD datasets.
% \item Most similar trajectory search: we evaluated the model using Top-1 Hit Rate (HR@1), Top-5 Hit Rate (HR@5), and Top-10 Hit Rate (HR@10), where Top-k Hit rate indicates the probability that the ground truth is in the top-k most similar samples ranked by the model.
% \item Metrics used in traffic state tasks are: MAE, MAPE, RMSE.
% \item Trajectory Recovery: We evaluated our model on three types of mask ratios: $85\%$, $90\%$, $95\%$. The evaluation metric is the recovery accuracy and Macro-F1 on masked road segments.
% \end{itemize}
For trajectory Travel Time Estimation: we adopt three metrics, including mean absolute error (MAE), mean absolute percentage error (MAPE), and root mean square error (RMSE); For next hop prediction: Following the settings in~\cite{START}, we used three metrics: Accuracy (ACC), mean reciprocal rank at 5 (MRR@5) and Normalized Discounted Cumulative Gain at 5 (NDCG@5); For trajectory classification: We use Accuracy (ACC), F1-score (F1), and Area Under ROC (AUC) to evaluate binary classification on BJ dataset. Using Micro-F1, Macro-F1 and Macro-Recall on XA and CD datasets; For most similar trajectory search: we evaluated the model using Top-1 Hit Rate (HR@1), Top-5 Hit Rate (HR@5), and Top-10 Hit Rate (HR@10), where Top-k Hit rate indicates the probability that the ground truth is in the top-k most similar samples ranked by the model; Metrics used in traffic state tasks are: MAE, MAPE, RMSE; For trajectory Recovery: We evaluated our model on three types of mask ratios: $85\%$, $90\%$, $95\%$. The evaluation metric is the recovery accuracy and Macro-F1 on masked road segments.

\paratitle{Implementation Details.} We chose GPT-2 as our backbone, and conducted experiments on Ubuntu 20.04 using 4 NVIDIA H800 80GB GPUs with pytorch 1.8.1.
% In the first training stage, we employ the Adam optimizer~\cite{loshchilov2017decoupled} with a peak learning rate of 2e-4 and $5\%$ warm-up cosine scheduler. In the second stage, we set the the peak learning rate as 1e-5.
% In both stage, the batch size is 128, and the max input length is 512 in BJ, 256 in XA and CD.
The implementation of our model can be find in code~\footnote{\url{https://github.com/bigscity/BIGCity}}.

\subsection{Performance Comparison}
We conducted extensive experiments on all 3 datasets across 8 ST tasks. As current baselines handle trajectories and traffic states separately, we compared \name to trajectory and traffic state baselines individually. All comparison experiments are repeated ten times, we list the mean values in this paper and provide standard deviation in the code link.

\paratitle{Trajectory Tasks.} Firstly, we conclude {\em trajectory next-hop prediction (Next), travel time estimation (TTE),  trajectory classification (CLAS)}, and {\em similar trajectory search (Simi)} as trajectory-based non-generative tasks as these four tasks share similar pipelines. Notably, current baselines require individual training on each task, while \name handles these tasks with a single set of parameters. Following the settings in~\cite{ma2024more}, {\em Next} task prediction the next token of an input, In {\em TTE}, we mask off all timestamps, and the ground truth for each segment is the time interval between its former one segment. {\em Simi} takes a certain input samples as queries and search the most similar samples from the whole dataset, and more details please refer to~\cite{ma2024more}. {\em CLAS} in \bj is a binary classification. In \xa and \cd, {\em CLAS} is a user-trajectory link classification and we only retained users with more than 50 trajectories. The comparison results are given in Tab.~\ref{tab:TBNG_task}. \name consistently and significantly outperforms the baselines across all scenarios.
% 0.585\textcolor{red}{with $xx\%|xx\%$ for the average and the largest improvement.}

Subsequently, {\em trajectory recovery (Reco)} usually employs task-specific models. For each input data, we sample it at a specific frequency to obtain a low-frequency trajectory and use the original trajectory as the target for trajectory recovery. We evaluate model’s performance under different
mask ratios. The comparison results are given in Tab.~\ref{tab:traj_recovery}.

\begin{table}[t]
\renewcommand{\arraystretch}{1.1}
\captionsetup{font=small}
\tabcolsep=0.12cm
  \centering
  \caption{Generalization experiments on three tasks, \ie travel time estimation, next hop prediction (Next), trajectory classification (CLAS)}
    \begin{tabular}{c|c|cc|cc|cc}
    \toprule[0.7pt]
    \multirow{2}{*}{\bf{Data}} & \multirow{2}{*}{Tasks} & \multicolumn{2}{c|}{{\bf TTE}} & \multicolumn{2}{c|}{{\bf Next}} & \multicolumn{2}{c}{{\bf CLAS}} \\
&  & MAE$\downarrow$   & RMSE$\downarrow$  & ACC$\uparrow$   & MRR@5$\uparrow$  & Mi-F1$\uparrow$ & Ma-F1$\uparrow$ \\ \midrule
\midrule
    \multirow{3}{*}{\rotatebox{90}{\bf XA}} & \Name  & 1.72  & 2.61  & 0.837  & 0.923 & 0.110 & 0.104 \\
& \Sname-BJ & 1.82  & 2.78  & 0.806  & 0.912 & 0.103 & 0.097  \\
\rowcolor{gray!10} & \textbf{Loss}  & 5.89\% & 5.98\% & 3.81\% & 1.20\% & 6.46\% & 6.94\% \\
\midrule
\multirow{3}{*}{\rotatebox{90}{\bf CD}} & \Name  & 1.29  & 2.18  & 0.821  & 0.910  & 0.153 & 0.169  \\
& \Sname-BJ & 1.37  & 2.31  & 0.792  & 0.878  & 0.144 & 0.159 \\
\rowcolor{gray!10} & \textbf{Loss}  & 5.63\% & 5.72\% & 3.62\% & 1.12\% & 6.22\% & 6.15\% \\
    \bottomrule[0.7pt]
    \end{tabular}%
  \label{tab:transfer}%
  \vspace{-0.3cm}
\end{table}%

\paratitle{Traffic State Tasks.} We evaluate \name on three traffic state tasks, \ie {\em one-step prediction (O-Step)}, {\em multi-step prediction (M-Step)}, and {\em traffic state imputation (TSI)}. Specifically, one-step prediction generates the traffic state in the next time slice, and multi-step prediction generates traffic states in the next 6 time slice. In traffic state imputation, we masked $25\%$ input data and trained models to recovery the masked data. Here, we only conducted experiments on \xa and \cd, as the trajectories in \bj are too sparse to get reliable traffic states. As shown in Tab.~\ref{tab:traffic_prediction}, our model largely outperforms the other baselines for both one-step and multi-step predictions on the two datasets. The average performance advantage is $\bm{2.83}\%$, with the largest reaching $\bm{7}\%$. As for traffic state imputation, the average and largest improvement are $\bm{7.8}\%|\bm{12.3}\%$.

\vspace{-0.1cm}
In above experiments, \name outperforms 17 baselines (7 representation models and 10 task-specific models) across all tasks, datasets, and metrics, indicating its universality and robust multi-task ability.

\subsection{Generalization Ability}
We evaluate \name's generalization ability in across datasets scenario.
Specifically, we first trained the whole \name on \bj dataset. Then, we transfer BIG-BJ's backbone model to \xa and \cd dataset. Specifically, we combined the spatiotemporal tokenizer of \xa and \cd with the backbone trained on \bj, and only fine-tuned the last MLP layer of tokenizers on XA and CD datasets.
The experimental results are given in Tab.~\ref{tab:transfer}.
In both \xa and \cd, \name-BJ denotes the transferred model with its backbone trained on \bj, and \name is the model completely trained on \xa or \cd.
The generalization ability is evaluated by the performance loss between \name and \Sname-BJ.
As shown in the table, the average performance degradation for \name-BJ compared with original \name model is within $\bm{7}\%$. In comparison with the results in Tab.~\ref{tab:TBNG_task}, \name-BJ still outperforms all of baselines in most cases, demonstrating our \name's superior cross-city generalization capability as well as the robustness. The experimental results indicate that our model has potential to be applied in a valuable scenario: pre-training the backbone model on a large city with ample data and then transferring it to smaller cities with limited data.

\subsection{Ablation Studies}
We consider the superior performance of \name stems from two key factors: $\bf{1)}$ The general representation captures comprehensive ST features, and $\bf{2)}$ the co-training mechanism in task-oriented tuning enhances ST feature sharing across tasks. In this section, we conduct analysis on the above hypotheses and evaluate the contributions of each module. Specifically, we first conducted ablations on model designs, followed by ST co-training mechanism. All experiments are conducted on the \xa dataset.
{\em Notably, some modules are essential to certain tasks. For example, there won't be trajectory tasks if the static encoder is absent. In Tab.~\ref{tab:ablation_tokenizer} and Tab.~\ref{tab:ablation_ST_Tasks}, $'-'$ denotes such tasks.}

\paratitle{Ablations on Model Designs.} We first conducted ablations on the static encoder, dynamic encoder of the spatiotemporal tokenizer, thereby evaluating the contribution of general representation. Subsequently, we conduct ablations on task-oriented prompts. Specifically, $\bf{1}$) {\em w/o-Dyn} removes the dynamic encoder, resulting in the loss of dynamic traffic state information. $\bf{2}$) {\em w/o-Sta} removes the static encoder, resulting in the loss of static topology information. $\bf{3}$) {\em w/o-Fus} removes the fusion encoder, resulting in the loss of long-range ST features integration. $\bf{4)} ${\em w/o-Pro} removes prompts from the input, and we trained a specific task MLP for each individual task. As shown in Tab.~\ref{tab:ablation_tokenizer}, the fusion encoder benefits all ST tasks, especially in tasks prefer long-range ST dependencies, \ie travel time estimation. Results in {\em w/o-Sta} and {\em w/o-Dyn} indicate that the general representation indeed enhances \name's performance. Dynamic features enhance trajectory tasks by improving representational distinctiveness, while static features benefit traffic state tasks by incorporating long-range topology information through input trajectories. As for task-oriented prompts, they have the most significant influence on performance of all tasks.The experimental results demonstrate their importance in \name's multi-task ability, as they provide task-specific guidance.

\paratitle{Ablations on Training.} We consider multi-task co-training benefits \name in feature exchanges among tasks. Therefore, we conduct ablations on threehighly heterogeneous tasks, \ie next hop prediction ({\em Next}), travel time estimation ({\em TTE}), multi-step traffic state prediction ({\em MS}). {\em All} denotes co-trained all three tasks. We incrementally add the task type for multi-task co-training. Experimental results in Tab.~\ref{tab:ablation_ST_Tasks} indicate that: the greater the differences in task types and data features (multi-step prediction and next hop prediction), the more significant the model's gains from multi-task co-training.

\begin{table}[t]
\fontsize{7.2pt}{7.5pt}\selectfont
\renewcommand{\arraystretch}{1.25}
\captionsetup{font=small}
\tabcolsep=0.07cm
  \centering
  \caption{Ablation Studies on Model Designs}
    \begin{tabular}{c|ccccccc|c}
    \toprule
    \multirow{2}{*}{\bf{Task}} & {TTE} & CLAS  & Next  & {Simi.} & {Reco.} & {TSI} & {M-Step} & \multirow{2}{*}{AveGAP} \\
    & MAE↓  &  Ma-F1 ↑ &  Acc ↑  & HR10 ↑  & Acc. ↑  & MAPE↓  & MAPE↓  &  \\
    \midrule
    \midrule
    w/o-Dyn+Fus & 1.96	& 0.102 & 0.803 & 0.801 & 0.537 & -  & -  & - \\
    \rowcolor{gray!10} GAP   & 13.9\% & 2.2\% & 4.1\% & 11.8\% & 4.4\% & - & - & 7.28 \\
    \midrule
    w/o-Dyn & 1.87  & 0.102 & 0.815 & 0.805 & 0.550 & -     & -     & - \\
    \rowcolor{gray!10} GAP   & 11.7\% & 2.2\% & 2.7\% & 11.4\% & 2.2\% & -     & -     & 6.9\% \\
    \midrule
    w/o-Sta+Fus & -  & - & - & - & - & 7.18  & 14.83  & - \\
    \rowcolor{gray!10} GAP   & - & - & - & - & - & 6.8\% & 4.9\% & 5.8\% \\
    \midrule
    w/o-Sta & -     & -     & -     & -     & -    & 7.02  & 14.48  &  - \\
    \rowcolor{gray!10} GAP   & - & - & - & - & -    & 5.3\% & 3.4\% & 4.4\% \\
    \midrule
    w/o-Pro & 2.03  & 0.096 & 0.759 & 0.892 & 0.513 & 7.672  & 15.51  & - \\
    \rowcolor{gray!10} GAP   & 18.0\% & 11.5\% & 9.3\% & 1.8\% & 8.7\% & 15.0\% & 9.6\% & 10.5\% \\
    \midrule
    \bf{\name} & 1.72   & 0.104 & 0.837 & 0.909 & 0.562 & 6.671  & 14.01 & - \\
    \bottomrule
    \end{tabular}%
  \label{tab:ablation_tokenizer}%
\end{table}%

\begin{table}[t]
\fontsize{8pt}{8pt}\selectfont
\renewcommand{\arraystretch}{0.75}
\captionsetup{font=small}
\tabcolsep=0.16cm
  \centering
  \caption{Ablation Studies on ST Tasks. MS denotes the multi-step traffic state prediction tasks}
    \begin{tabular}{c|c|c|c|c|c|>{\columncolor{gray!10}} c}
    \toprule
    \multirow{2}[4]{*}{\bf{Metric}} & \multicolumn{6}{c}{\bf{Task}} \\
    \cmidrule{2-7}          & Next  & TTE   & MS & MS+Next & TTE+Next & \bf{All} \\
    %\midrule
    \midrule
    ACC $\uparrow$  & 0.79  & -     & -     & 0.82  & 0.8   & \textcolor{black}{\bf{0.84}} \\
    \midrule
    MAE $\downarrow$  & -     & 1.8   & -     & -     & 1.77  & \textcolor{black}{\bf{1.75}} \\
    \midrule
    MAPE $\downarrow$ & -     & -    & 14.36 & 14.19 & -     & \textcolor{black}{\bf{14.14}} \\
    \bottomrule
    \end{tabular}%
  \label{tab:ablation_ST_Tasks}%
  % \vspace{-0.2cm}
\end{table}% 

%% file: content/5_model_analysis.tex
\vspace{0.2cm}
\section{Model Analysis}
% \vspace{-0.2cm}
\subsection{Parameter Sensitivity}
We conducted parameter sensitivity analysis for critical hyper-parameters, \ie the quantity of LoRA modules as well as the matrix rank within each module. The rate of LoRA modules are denoted as $n$, and the matrix rank in each LoRA module is denoted as $r$. We set $n$ to $1$, $1/2$, and $1/3$. For example, $n= 1/2$ means the there are $50\%$ of GPT-2’s attention blocks are attached with LoRA modules. The range of $r$ is $4$, $8$, $16$, $32$.

\begin{figure}[t]
\captionsetup{font=small}
    \centering
    \captionsetup[subfigure]{labelformat=simple}
    \begin{subfigure}[b]{0.15\textwidth}
        \includegraphics[width=\textwidth]{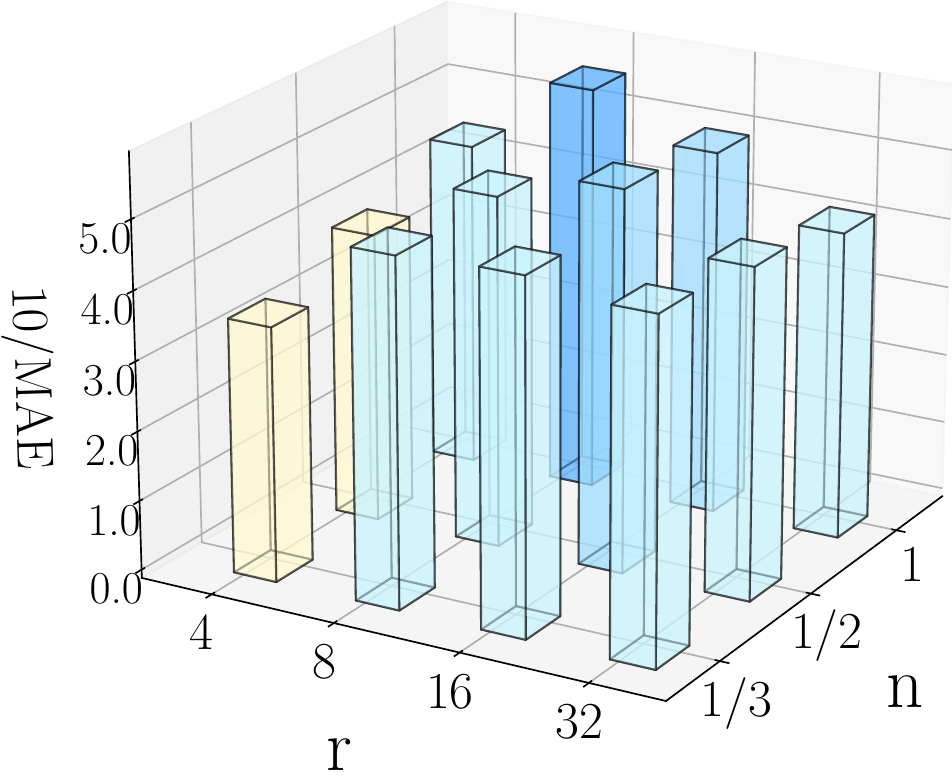}
        % \vspace{-0.3cm}
        \caption{ \scriptsize MAE on travel time estimation task}
        \label{sen:ETA_MAE}
    \end{subfigure}
    \hfill
    \begin{subfigure}[b]{0.15\textwidth}
        \includegraphics[width=\textwidth]{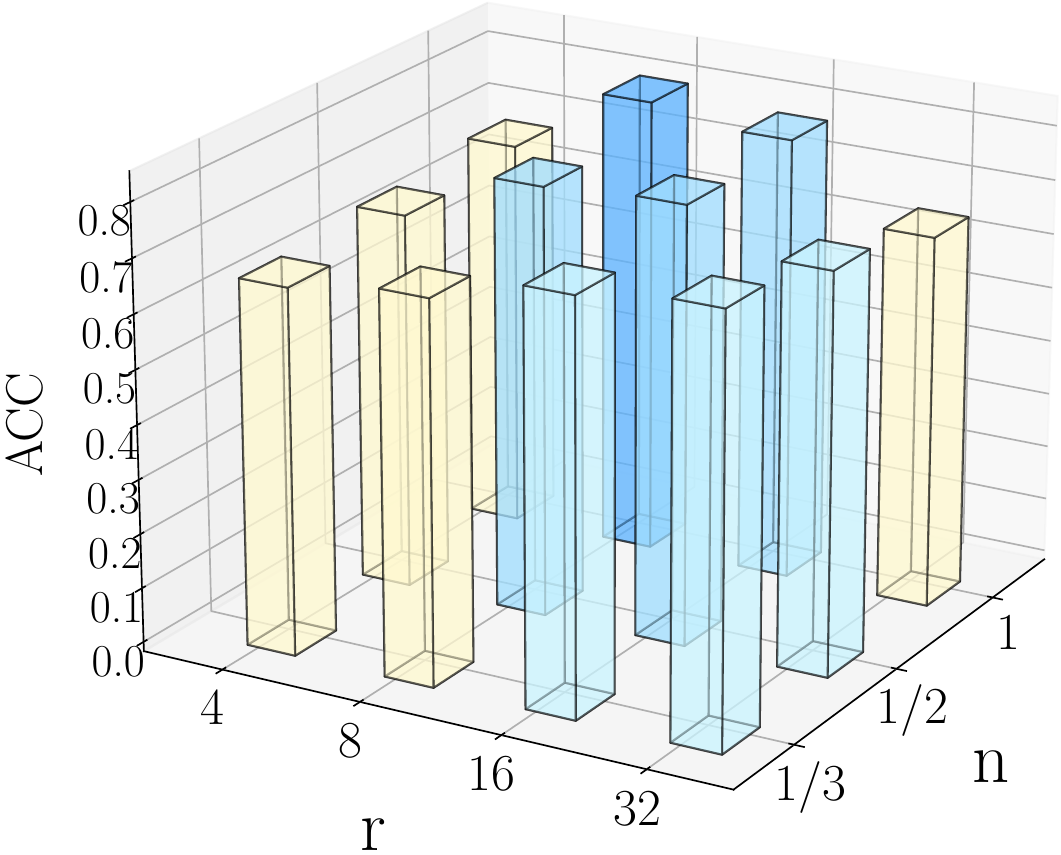}
        % \vspace{-0.3cm}
        \caption{ \scriptsize Accuracy on next hop prediction task}
        \label{sen:NexH_ACC}
    \end{subfigure}
    \hfill
    \begin{subfigure}[b]{0.15\textwidth}
        \includegraphics[width=\textwidth]{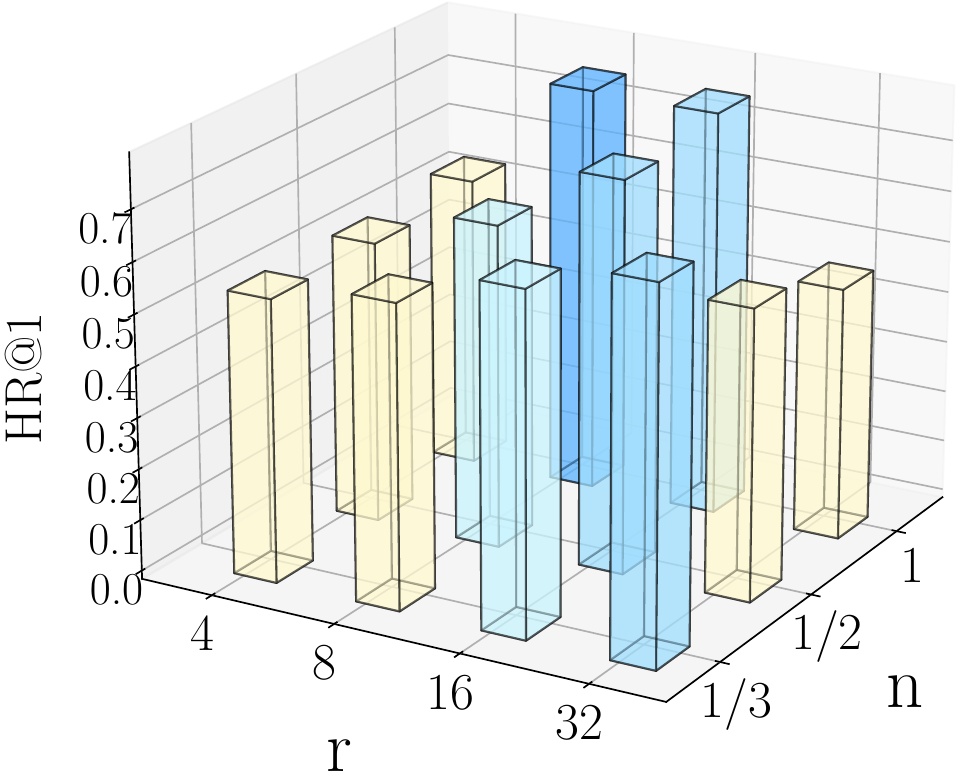}
        % \vspace{-0.3cm}
        \caption{\scriptsize HR@1 on most similar search task}
        \label{sen:Most_HR1}
    \end{subfigure}
    \\
    \begin{subfigure}[b]{0.15\textwidth}
        \includegraphics[width=\textwidth]{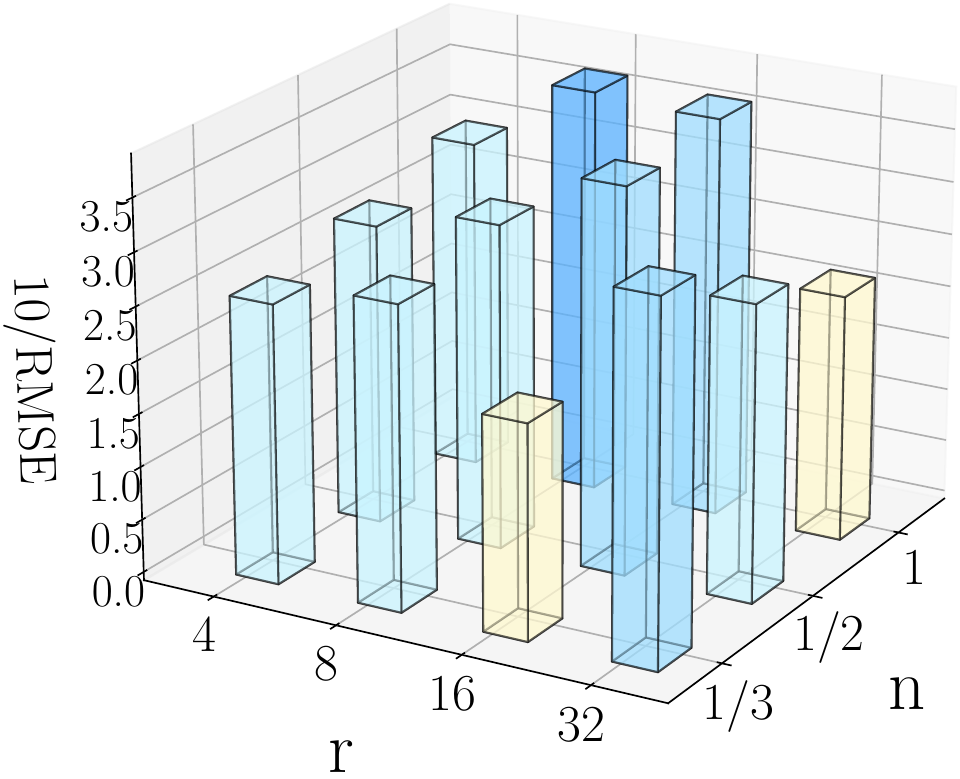}
        % \vspace{-0.3cm}
        \caption{\scriptsize RMSE on travel time estimation task}
        \label{sen:ETA_RMSE}
    \end{subfigure}
    \hfill
    \begin{subfigure}[b]{0.15\textwidth}
        \includegraphics[width=\textwidth]{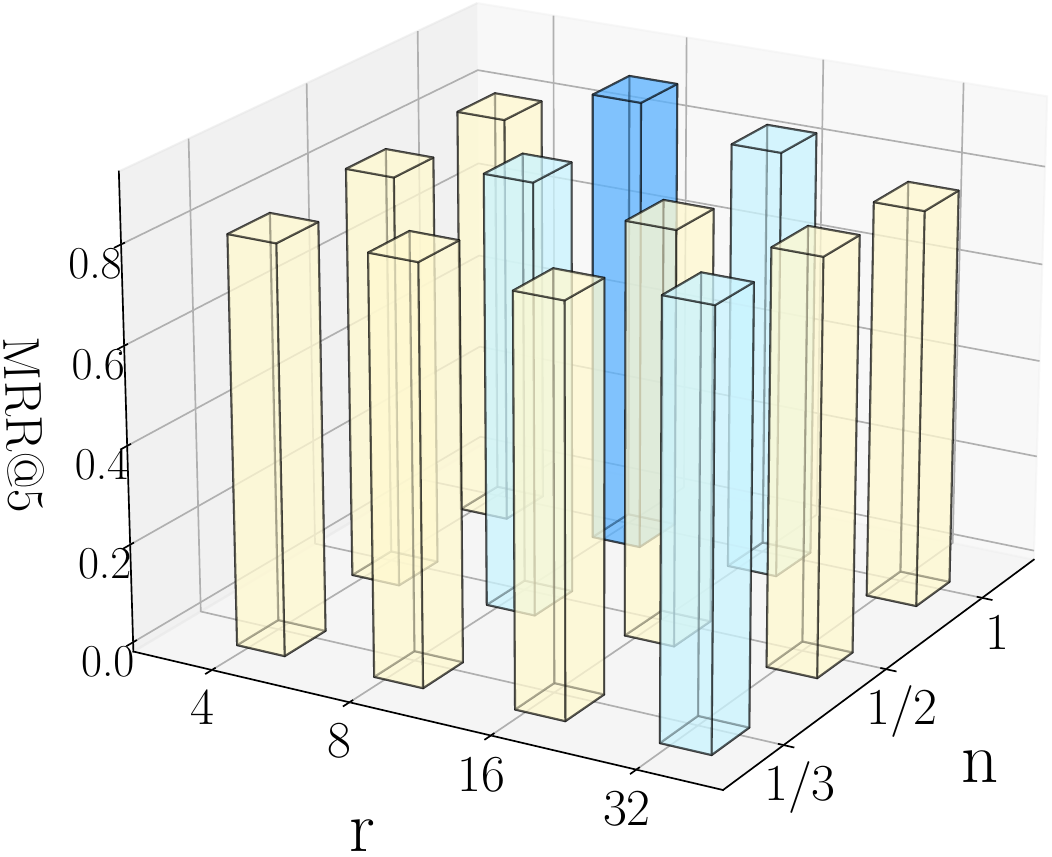}
        % \vspace{-0.3cm}
        \caption{\scriptsize MRR@5 on next hop prediction task}
        % \label{sen:NexH_MRR}
    \end{subfigure}
    \hfill
    \begin{subfigure}[b]{0.15\textwidth}
         \includegraphics[width=\textwidth]{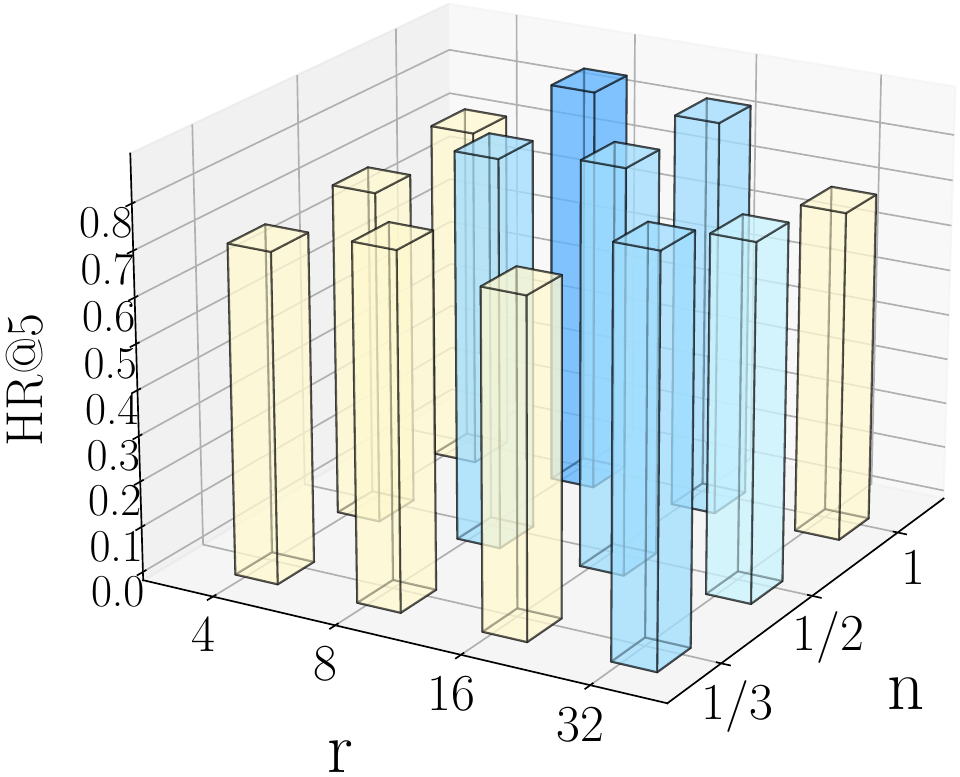}
        % \vspace{-0.3cm}
        \caption{\scriptsize HR@5 on most similar search task}
        \label{sen:Most_HR5}
    \end{subfigure}
    % \vspace{0.2cm}
    \caption{The parameter sensitivity analysis on classification, regression, and comparison task}
    % \vspace{-0.2cm}
    \label{fig:test}
\end{figure}

Specifically, we conduct analysis on three types of tasks, \ie regression task (travel time estimation), classification task (next hop prediction), comparison task (most similar search).
The results of travel time estimation are presented in Fig.~\ref{sen:ETA_MAE} and Fig.~\ref{sen:ETA_RMSE}. Since smaller MAE and MAPE means better model performance, we use $\frac{10}{MAE}$ and $\frac{10}{RMSE}$ on the vertical axis to represent MAE and MAPE, respectively. The results in next hop prediction are in Fig.~\ref{sen:NexH_ACC} and Fig.~\ref{sen:Most_HR5}, where Accuracy and MRR@5 are metrics. The results in most similar trajectory search are in Fig.~\ref{sen:Most_HR1} and Fig.~\ref{sen:Most_HR5}, where HR@1 and HR@5 are metrics. The experimental results show that model's performance improves with increasing $n$. However, \name is more sensitive to $r$. When $r \le 8$, the model's performance increases with $r$, but for $r \ge 16$, the performance deteriorates as $r$ continues to rise. Considering both computational cost and performance, we selected $n = 1$ and $r = 8$.

\vspace{0.2cm}
\subsection{Model Efficiency And Scalability}
\begin{figure}[t]
    \centering
    \captionsetup[subfigure]{labelformat=simple, font=tiny}
    \captionsetup{font=small}
    \begin{subfigure}{0.155\textwidth}
        \includegraphics[width=\textwidth]{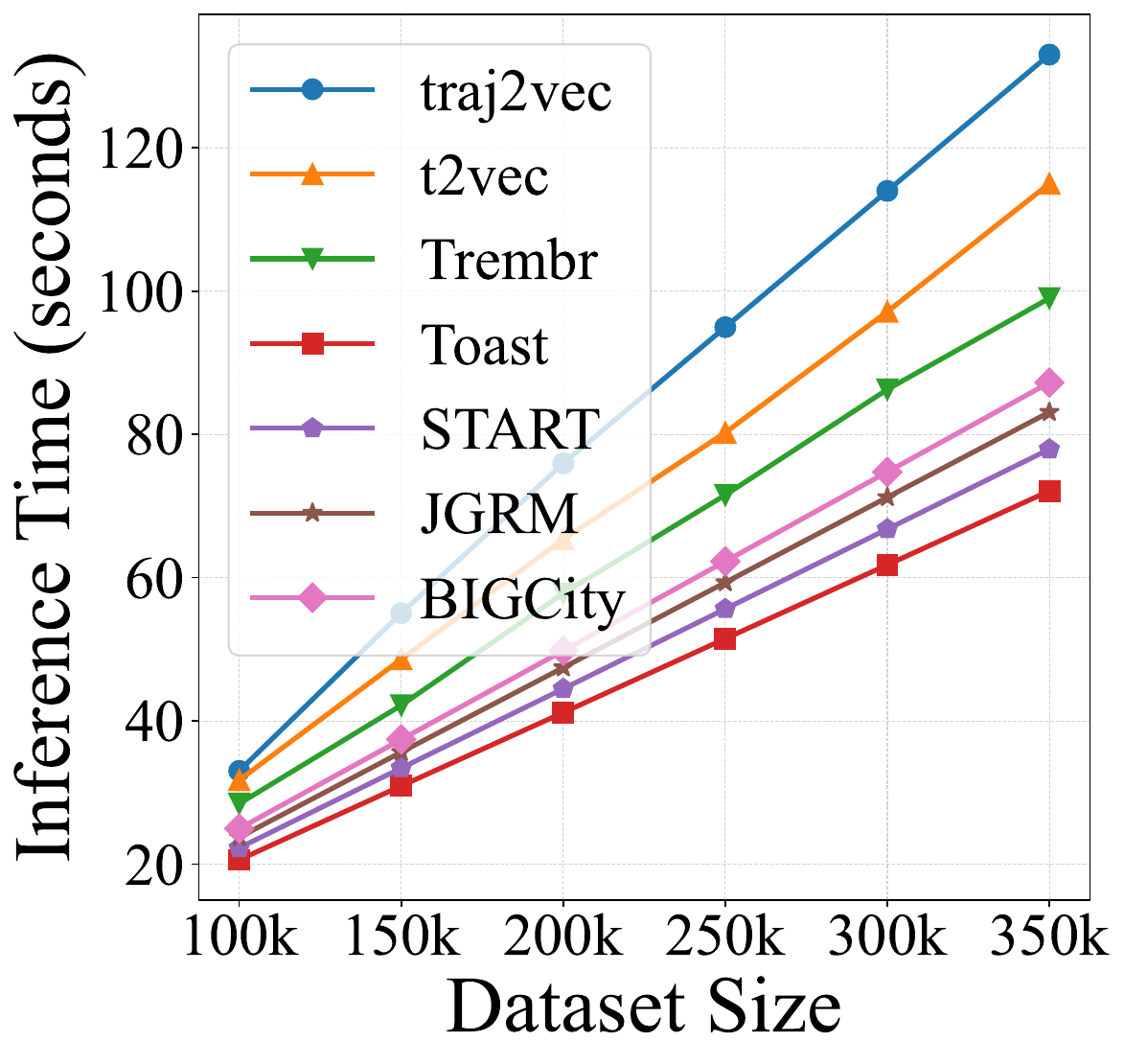}
        % \vspace{-0.5cm}
        \caption{Efficiency: Inference Time}
        \label{sen:infernce}
    \end{subfigure}
    \begin{subfigure}{0.161\textwidth}
        \includegraphics[width=\textwidth]{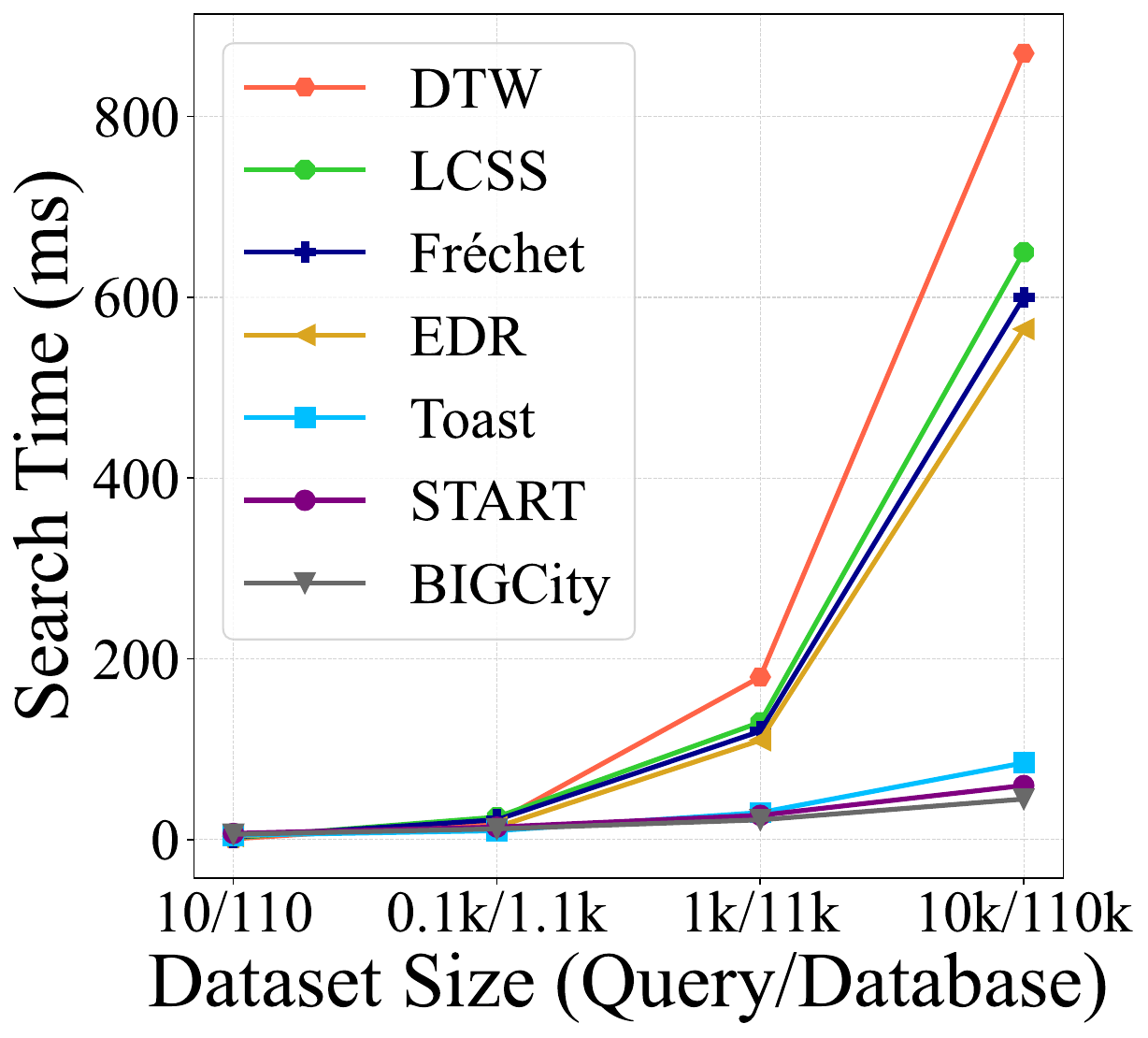}
        % \vspace{-0.5cm}
        \caption{Scalability: Search Time}
        \label{sca:time}
    \end{subfigure}
    \begin{subfigure}{0.16\textwidth}
        \includegraphics[width=\textwidth]{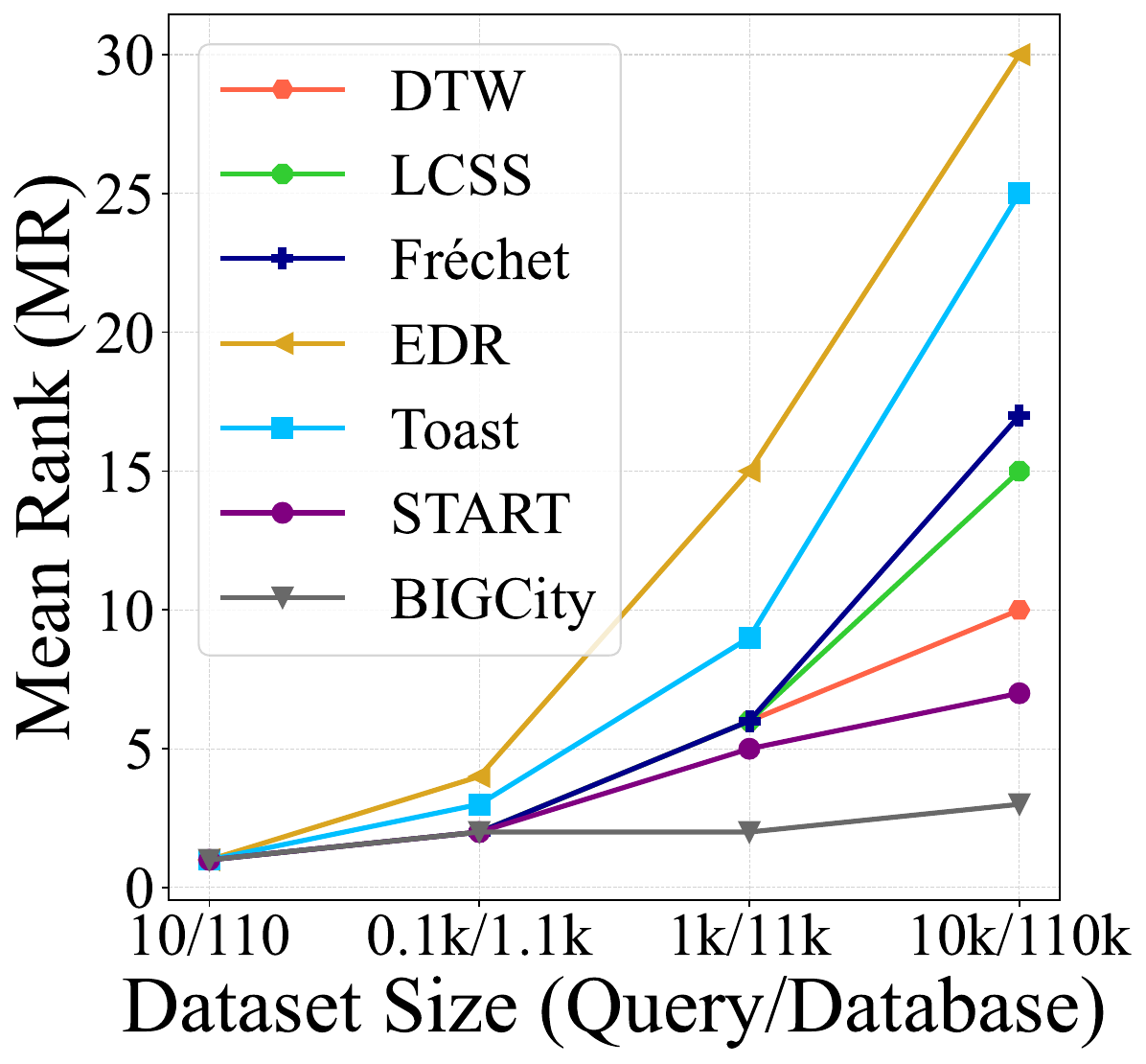}
        % \vspace{-0.5cm}
        \caption{scability: Performace}
        \label{sca:performance}
    \end{subfigure}
    \caption{ (a) Inference Efficiency: Time required to generate ST representations as data size grows. (b) Average Search Time: Time taken for query searches as data size grows. (c) Mean Rank: Average rank of ground truth in similarity-based sorting; lower is better.}
    \label{fig:efficiency_scalability}
    % \vspace{-0.2cm}
\end{figure}

We conducted efficiency and scalability experiments on thr \xa dataset, with the number of input samples increase from $100$k to $350$k. The experimental results are shown in Fig.~\ref{fig:efficiency_scalability}. Similar trends are observed in other datasets.

\begin{table}[t]
\fontsize{8.5pt}{9pt}\selectfont
\renewcommand{\arraystretch}{1.2}
\captionsetup{font=small}
  \centering
  \caption{The Efficiency Analysis on \xa. Stage-1 is the representation training, Stage-2 is the task-related tuning}
    \begin{tabular}{c|c|c|c}
    \toprule
    \textbf{Models}  & \begin{tabular}[c]{@{}c@{}}\bf{GPU Usage}\\ (GB)\end{tabular} & \begin{tabular}[c]{@{}c@{}}{\bf Stage-1 Speed}\\ (min/epoch)\end{tabular} & \begin{tabular}[c]{@{}c@{}}{\bf Stage-2 Speed}\\ (min/epoch)\end{tabular} \\
    \midrule
    \midrule
    Traj2vec & 4.932 & 4.662 & 3.761 \\
    Toast & 5.271 & 4.801 & 3.996 \\
    START & 21.65 & 14.93 & 7.753 \\
    \rowcolor{gray!10}\bf{\name} & 28.73 & 11.32 & 8.681 \\
    \bottomrule
    \end{tabular}%
  \label{tab:efficiency_table}%
  % \vspace{-.2cm}
\end{table}%

\paratitle{Efficiency.} Although with the largest parameter size, \name still demonstrates the efficiency advantage as a multi-task model in both training and inference time cost.

For training, we compare \name with those two-stage training baselines on travel time estimation task. As a two-stage training model, we consider training cost of both stage 1 (representation training) and stage 2 (task tuning). We provide GPU usage, Training speed in stage 1, and stage 2. As shown in Tab.~\ref{tab:efficiency_table}, \name has the largest parameter size but with moderate training cost. Even in stage 1, \name is faster than the smaller model like START. This is because we trained \name by LoRA, ensuring that there is a reasonable trade-off between additional learnable parameters and training speed.

For inference, we record the time cost of representing $100$k–$350$k input samples by the backbone. As shown in Fig~\ref{sen:infernce}, \name outperform RNN models in speed while having the speed similar to models with three times fewer parameters~\cite{START}. \name has the largest parameters yet maintains moderate inference cost. This is because: $\bm{1)}$ The RNN model operates sequentially, whereas the attention mechanism in \name is parallel.
$\bm{2)}$ The causal attention mechanism in \name has lower computational cost than self-attention, yet most baselines employ self-attention.

\paratitle{Scalability.}
In the sequel, we explore the scalability performance of \name compared with six baselines, focusing on their ability to handle variations in dataset size. We conducted experiments on the most similar trajectory search task. It is suitable to demonstrate model scalability, as the volume of input data has significant influence on model's performance and efficiency. We conduct experiments on inference time and performance in Fig~\ref{sca:time} and Fig.~\ref{sca:performance}. Specifically, the size of query sample varies from 10 to 10000, and the size of overall data is ten times of query samples. Besides SOTA deep learning baselines, we also involve some traditional algorithm. including Dynamic Time Warping (DTW)~\cite{yi1998efficient}, Longest Common SubSequence (LCSS)~\cite{vlachos2002discovering}, Fr’echet Distance~\cite{alt1995computing}, and Edit Distance on Real Sequence (EDR)~\cite{chen2005robust}. As shown in Fig.~\ref{sca:time}, the inference cost of \name linearly associates with the data volume but traditional methods is largely affected by data size. On the other hand, as shown in Fig.~\ref{sca:performance}, \name maintains robust performance with increasing data size. Unlike \name, other baselines exhibit significant performance degradation. In summary, \name is robust in performance and the computational cost linearly associates with the data volume. The experimental results indicate \name has great potential in scaling to large datasets.

% To further evaluate data utilization efficiency of our model, we conducted transfer experiments on both \name and START. Following the settings in Section 4, we trained them on three tasks using 20\%, 40\%, 60\%, and 100\% of the \xa dataset. The results, shown in Figures 1, 2, and 3, demonstrate that \name consistently outperform START, achieving better performance with less data. More importantly, \name consistently benefits from the increasing data volume. This indicates that \name requires lower training costs to achieve comparable results, and has high data utilization efficiency.

% \begin{figure}[t]
%     \centering
%     \includegraphics[width=0.9\linewidth]{figure/prompt_templates/Next Hop prediction.pdf}
%     \caption{The prompt template of next hop prediction task}
%     \label{next_hop_template}
% \end{figure}

% \begin{figure}[t]
%     \centering
%     \includegraphics[width=0.9\linewidth]{figure/prompt_templates/ETA.pdf}
%     \caption{The prompt template of trajectory travel time estimation task}
%     \label{ETA}
% \end{figure}

% \begin{figure}[t]
%     \centering
%     \includegraphics[width=0.9\linewidth]{figure/prompt_templates/Traffic State.pdf}
%     \caption{The prompt template of traffic state prediction task}
%     \label{traffic_state}
% \end{figure}

% \begin{figure}[t]
%     \centering
%     \includegraphics[width=0.9\linewidth]{figure/prompt_templates/Trajectory Recovery.pdf}
%     \caption{The prompt template of trajectory recovery task}
%     \label{traj_recovery}
% \end{figure}